\documentclass{article}
\usepackage{arxiv}  

\usepackage[utf8]{inputenc}
\usepackage{subcaption}
\usepackage{booktabs}
\usepackage{bm}
\usepackage{bbm}
\usepackage{tcolorbox}
\usepackage{multicol,multirow}
\usepackage{natbib}
\usepackage{comment}
\usepackage{booktabs}

\usepackage{enumerate}   
\usepackage{amsmath}
\usepackage{tikz}

\usetikzlibrary{positioning,arrows}

\usepackage{wrapfig}
\DeclareMathOperator*{\argmax}{arg\,max}
\DeclareMathOperator*{\argmin}{arg\,min}

\usepackage{amsmath}
\usepackage{appendix}
\usepackage{amssymb}

\usepackage{amsthm}
\usepackage[capitalise]{cleveref}
\Crefname{assumption}{Assumption}{Assumptions}

\theoremstyle{plain}
\newtheorem{theorem}{Theorem}

\newtheorem{corollary}{Corollary}
\theoremstyle{definition}
\newtheorem{assumption}{Assumption}
\newtheorem{proposition}{Proposition}
\newtheorem{definition}{Definition}
\newtheorem{remark}{Remark}

\newcommand{\iid}{ \stackrel{\mathrm{i.i.d.}}{\sim} }

\def\Cramer{Cram\'{e}r}

\newcommand{\E}{\mathbb{E}}

\newcommand{\pa}{\mathrm{\pa}}

\newcommand{\epol}{\pi^\mathrm{e}}
\newcommand{\bpol}{\pi^\mathrm{b}}

\newcommand{\es}{\mathrm{e}}
\newcommand{\bs}{\mathrm{b}}

\renewcommand{\eqref}[1]{(\ref{#1})}

\newcommand{\RN}[1]{%
  \textup{\uppercase\expandafter{\romannumeral#1}}%
}

\usepackage{color,lscape,longtable}

\def\boxit#1{\vbox{\hrule\hbox{\vrule\kern6pt\vbox{\kern6pt#1\kern6pt}\kern6pt\vrule}\hrule}}

\usepackage{xcolor}
\newcount\Comments  
\newcommand{\kibitz}[2]{\ifnum\Comments=1\textcolor{#1}{#2}\fi}

\usepackage{algorithm}
\usepackage{algorithmic}
\usepackage{braket}
\usepackage{mathtools}

\usepackage[switch]{lineno}

\usepackage{times}  
\usepackage{helvet} 
\usepackage{courier}  
\usepackage[hyphens]{url}  
\usepackage{graphicx} 
\usepackage{natbib}  
\usepackage{caption} 

\usepackage{color}

\title{A Practical Guide of Off-Policy Evaluation for Bandit Problems}

\newcommand*{\affaddr}[1]{#1} 
\newcommand*{\affmark}[1][*]{\textsuperscript{#1}}
\newcommand*{\equalcontribution}[1][*]{\textsuperscript{*}}
\newcommand*{\email}[1]{\texttt{#1}}

\author{%
Masahiro Kato\affmark[1],\ \ \ Kenshi Abe\affmark[1]\footnotemark[1]\thanks{Equal contributions.},\ \ \ Kaito Ariu\affmark[2]\footnotemark[1],\ \ \ Shota Yasui\affmark[1]\\
\affaddr{\affmark[1]CyberAgent Inc.}\\
\email{masahiro\_kato@cyberagent.co.jp}\\
\email{abe\_kenshi@cyberagent.co.jp}\\
\email{yasui\_shota@cyberagent.co.jp}\\
\affaddr{\affmark[2] KTH}\\
\email{ariu@kth.se}\\
}

\begin{document}

\maketitle

\begin{abstract}
\emph{Off-policy evaluation} (OPE) is the problem of estimating the value of a target policy from samples obtained via different policies. Recently, applying OPE methods for \emph{bandit problems} has garnered attention. For the theoretical guarantees of an estimator of the policy value, the OPE methods require various conditions on the target policy and policy used for generating the samples. However, existing studies did not carefully discuss the practical situation where such conditions hold, and the gap between them remains. 
This paper aims to show new results for bridging the gap. Based on the properties of the evaluation policy, we categorize OPE situations. Then, among practical applications, we mainly discuss the \emph{best policy selection}. For the situation, we propose a meta-algorithm based on existing OPE estimators. We investigate the proposed concepts using synthetic and open real-world datasets in experiments.
\end{abstract}

\section{Introduction}
As an instance of sequential decision-making problems, the \emph{multi-armed bandit} (MAB) problem has attracted significant attention in various applications, such as ad optimization \citep{narita2019counterfactual}, personalized medicine \citep{ChowChang201112,Villar2018}, and recommendation systems \citep{Li2010,Li2016}. However, because we only observe an action chosen following a \emph{behavior policy}, which yields \emph{sample selection bias} in the observed dataset, it is not easy to evaluate a new policy (\emph{evaluation policy}) from the log data. For solving this problem, \emph{off-policy evaluation} (OPE) is proposed \citep{kdd2009_ads, Li2010, dudik2011doubly,Li2011,wang2017optimal,narita2019counterfactual,pmlr-v97-bibaut19a,Kallus2019IntrinsicallyES,Oberst2019}. 

As reviewed in this paper, OPE estimators require several conditions to achieve theoretical properties, such as \emph{unbiasedness}, \emph{consistency}, and \emph{asymptotic normality}. For instance, we need to put an assumption such that an evaluation policy is independent of log data used for constructing the OPE because the correlation between the evaluation policy and an OPE estimator causes a biased result \citep{Li2011,kato_uehara_2020,kallus2020stochasticpolicy}. However, although real-world applications often break the condition, existing studies did not pay much attention to which practical situation satisfies the condition.

In addition, we point out terminology problems in OPE. A confusing terminology may cause misuse of OPE methods. For example, \citet{narita2019counterfactual} and \citet{saito2020large} study methods for evaluating bandit policies, such as the Thompson sampling, but their experimental process has the potential to cause unexpected bias because the evaluation probability is not deterministic \citep{narita2019counterfactual} and fail to reproduce trajectories of the bandit algorithms \citep{saito2020large}. Although they propose OPE methods for bandit policy, their methods estimate a weighted average of expected reward over a fixed probability of choosing an action, not a data-dependent bandit policy; that is, the definition of evaluation policy differs between their theoretical results and experiments. Thus, the confusing use of terminologies potentially misleads the applications of OPE.

Five contributions are made: (i) summarizing the potential concern and limitations of OPE methods; (ii) organizing the OPE terminologies; (iii) categorizing situations in which OPE methods are applicable; (iv) showing some new results bridging existing studies and practical situations. This paper is organized as follows. In Section~\ref{sec:prob} and \ref{sec:prelim}, we formulate our OPE problem setting and review OPE methods with the attention to the potential theoretical concerns. In Section~\ref{sec:cat_eval}, based on cases of evaluation probabilities, we categorize the OPE methods. Built upon the categorization, we introduce a practical application of OPE in Section~\ref{sec:beps} and show the experimental results in Section~\ref{sec:exp}.

\section{Problem Setting}
\label{sec:prob}
In this section, we describe the problem setting of OPE.

\subsection{Data-Generating Process}
Let $A_t$ be an action in $\mathcal{A}=\{1,2,\dots,K\}$, $X_t$ be the \emph{covariate} observed by the decision maker when choosing an action, and $\mathcal{X}$ be the space of covariate. Let us denote a random variable of a reward at period $t$ as $Y_t=\sum^K_{a=1}\mathbbm{1}[A_t = a]Y_t(a)$, where $Y_t(a):\mathcal{A}\to\mathbb{R}$ is a potential outcome\footnote{We can express the reward without using the potential reward variable. See Appendix~B of \citet{kato_uehara_2020}}. This setting is also called \emph{bandit feedback}. Suppose that we have access to a dataset $\mathcal{S}_T= \big\{(X_t, A_t, Y_t)\big\}^{T}_{t=1}$ with the following data-generating process (DGP):

\begin{align}
\label{eq:DGP}
\big\{(X_t, A_t, Y_t)\big\}^{T}_{t=1}\sim p(x)p_t(a\mid x)p(y\mid a, x),
\end{align}
where $p(x)$ denotes the density of the covariate $X_t$, $p_t(a\mid x)$ denotes the probability of choosing an action $a$ conditioned on a covariate $x$ at period $t$, and $p(y\mid a, x)$ denotes the density of a reward $Y_t$ conditioned on an action $a$ and covariate $x$. We assume that $p(x)$ and $p(y\mid a, x)$ are invariant across periods, but $p_t(a\mid x)$ can take different values across periods. 

Let us call a function determines the probability $p_t(a\mid x)$ a \emph{behavior probability}. For example, \citet{narita2019counterfactual} uses a behavior probability defined as $\bpol_t:\mathcal{X}\to \Delta^K$, where $\Delta^K := \{(p_a) \in [0,1]^K\mid \sum^K_{a=1}p_a=1\}$. We also define a behavior policy as a system that determines the behavior probability. In MAB problems, the behavior policy can be considered as a function of the trajectory. In most existing studies, a behavior policy and probability are not distinguished, and both are called a behavior policy.

\subsection{Value of Evaluation Policy}
We consider estimating the value of an \emph{evaluation policy} using samples obtained under the behavior policy. Let a policy generating a probability of choosing an action $\epol:\mathcal{X}\to\Delta^K$ be an evaluation policy. Similar to the behavior probability, we also call $\epol$ as the evaluation probability. In most existing studies, a behavior/evaluation policy and probability are not clearly distinguished, and both are called a behavior/evaluation policy. However, distinguishing them is critical, as mentioned; if an evaluation policy depends on a trajectory generated from the evaluation policy itself, we need to reproduce the trajectory for evaluating the evaluation policy. For example, MAB algorithms carefully select an arm in each period and balance the trade-off between exploration and exploitation using self-generated trajectories to update the parameter. However, in OPE, we have a trajectory generated from a behavior policy, which is usually different from that of the evaluation policy and cannot know the true real-time behavior of the evaluation policy. We discuss such terminology-related problems in Appendix~\ref{appdx:term}.

The goal of OPE for an evaluation policy generating an evaluation probability $\pi^{\mathrm{e}}(a \mid x)$ is to estimate the expected reward from the evaluation probability $\pi^{\mathrm{e}}(a \mid x)$ defined as $R(\epol) := \mathbb{E}\left[\sum^K_{a=1}\epol(a \mid x)Y_t(a)\right]$. 

To identify the policy value $R(\epol)$, we assume overlaps of the distributions of policies and the boundedness of reward. 
\begin{assumption}\label{asm:overlap_pol}
There exists a constant $C_1$ such that $0\leq \frac{\epol(a\mid x)}{p_t(a\mid x)}\leq C_1$.
\end{assumption}
\begin{assumption}\label{asm:overlap_reward}
There exists a constant $C_2$ such that $|Y_t| \leq C_2$.
\end{assumption}
In addition to the above assumptions, we also assume that an evaluation probability $\epol$ is deterministic \citep{Kallus2019IntrinsicallyES,kato_uehara_2020}. For each situation, a deterministic evaluation probability has different meanings, such as a random variable independent from $\mathcal{S}_T$, random variable constructed converging a time-invariant function as $T\to\infty$, or not a random variable. Note that the deterministic evaluation probability does not mean that the evaluation probability $\epol$ chooses a certain action with probability $1$. This assumption mainly implies that we cannot evaluate an evaluation probability constructed from samples correlated with the dataset $\mathcal{S}$. In Section~\ref{sec:cat_eval}, we consider practical situations with deterministic evaluation policies.

Besides, let us also define the Average Treatment Effect (ATE) between actions $a$ and $a'$. Although breaking the definition of the evaluation probability, we can regard for all $ x\in \mathcal{X}$, $\pi^{\mathrm{\es, (1)}}(a\mid x)=1$ and $\pi^{\mathrm{\es, (2)}}(a'\mid x)=1$ as a deterministic evaluation probability in ATE estimation and apply existing OPE estimators to estimate the value. In this paper, when referring to OPE, we allow estimations of ATE to be contained in the target of OPE estimators.

{\bf Notations:} Let $\mathcal{S}_{t-1}\in\mathcal{M}_{t-1}$ be the history until $t-1$ period defined as $\mathcal{S}_{t-1}=\{X_{t-1}, A_{t-1}, Y_{t-1}, \ldots, X_{1}, A_1, Y_{1}\}$ with the space $\mathcal{M}_{t-1}$. Let us denote $\mathbb{E}[Y_t(a)\mid x]$ and $\mathrm{Var}(Y_t(a)\mid x)$ as $f^*(a, x)$ and $\nu^*(a, x)$, respectively. Let $\mathcal{F}$ be the classes of $f^*(a, x)$ and $\hat{f}_{t}(a, x\mid \mathcal{S}_{t-1})$ be an estimator of $f^*(a, x)$ constructed from $\mathcal{S}_{t-1}$. Let $\mathcal{N}(\mu, \mathrm{var})$ be the normal distribution with the mean $\mu$ and the variance $\mathrm{var}$. We often denote a behavior probability as $\bpol_t$ or $\bpol$, and an evaluation probability as $\epol_t$ or $\epol$. Additionally, let $\|\mu(X,A,Y)\|_2$ be $\E[\mu^2(X,A,Y)]^{1/2}$ for the function $\mu$.

\section{Preliminaries of OPE}
\label{sec:prelim}
We review three types of well-known estimators of $R(\pi^{\mathrm{e}})$. For simplicity, let us assume that the behavior probability $\bpol_t$ is known. In the case where the behavior probability $\bpol_t$ is unknown, we replace the behavior probability with its estimator. The first estimator is an inverse probability weighting (IPW) type estimator \citep{rubin87,hirano2003efficient,swaminathan15a} given by $\frac{1}{T}\sum^T_{t=1}\sum^K_{a=1}\epol(a\mid X_t)\Gamma^{\mathrm{IPW}}_t(a; \bpol_t)$,
where 
\begin{align*}
&\Gamma^{\mathrm{IPW}}_t(a; \pi) = \Gamma^{\mathrm{IPW}}(a; X_t, A_t, Y_t, \pi) = \frac{\mathbbm{1}[A_t = a]Y_t}{\pi(A_t\mid X_t)}.
\end{align*}
Even though this estimator is unbiased when the behavior probability is known, it suffers from high variance. The second estimator is a direct method (DM) type estimator $\frac{1}{T}\sum^T_{t=1}\sum^K_{a=1}\epol(a\mid X_t)\Gamma^{\mathrm{DM}}_t(a; \hat{f})$, where 
\begin{align*}
&\Gamma^{\mathrm{DM}}_t(a; f) = \Gamma^{\mathrm{DM}}(a; X_t, f) = f(a, X_t),
\end{align*}
and $\hat{f}(a, X_t)$ is an estimator of $f^*(a, X_t)$ \citep{HahnJinyong1998OtRo}. This estimator is known to be weak against model misspecification of $f^*(a, X_t)$. The third estimator is an augmented inverse probability (AIPW) type estimator $\frac{1}{T}\sum^T_{t=1}\sum^K_{a=1}\epol(a\mid X_t)\Gamma^{\mathrm{AIPW}}_t(a; \bpol_t, \hat{f})$, where
\begin{align*}
\Gamma^{\mathrm{AIPW}}_t(a; \pi, f) & = \Gamma^{\mathrm{AIPW}}(a; A_t, X_t, Y_t, \pi, f)\\
& = \frac{\mathbbm{1}[A_t = a](Y_t - f(a, X_t))}{\pi(A_t\mid X_t)} + f(a, X_t).\nonumber
\end{align*}
When replacing the behavior probability with its estimator, AIPW estimator is called doubly robust estimator \citep{robins94,ChernozhukovVictor2018Dmlf}.

When theoretically evaluating OPE estimators, we often check the unbiasedness, consistency, asymptotic normality, and concentration inequality. For the above estimators, unbiasedness and/or consistency are satisfied in general. On the other hand, there are various methods for guaranteeing the asymptotic normality of the above estimator for specific choices of $\bpol$ and $\hat{f}$. For brevity, we explain the asymptotic distribution of the AIPW and Adaptive AIPW  (A2IPW) \citep{Kato2020} estimators for the cases in which the behavior probability is fixed and updated through periods, respectively. First, we consider the case where $\forall t,\; \bpol_t(a\mid x)=\bpol(a\mid x)$ in the DGP \eqref{eq:DGP}. In this case, by using some methods for deriving the asymptotic normality, such as double/debiased machine learning (DDM) proposed by \citet{ChernozhukovVictor2018Dmlf}, we can derive the asymptotic distribution of AIPW type estimator. Let us define an AIPW estimator with an estimator $\hat{f}_T$ of $f^*$ as $\widehat{R}^{\mathrm{AIPW}}_T(\epol)=\frac{1}{T}\sum^T_{t=1}\sum^K_{a=1}\epol(a| X_t)\Gamma^{\mathrm{AIPW}}_t(a; \bpol_t, \hat{f}_T)$. Then, under some conditions, we can show that $\sqrt{T}\left(\widehat{R}^{\mathrm{AIPW}}_T(\epol) - R(\epol)\right) \xrightarrow{\mathrm{d}}\mathcal{N}(0, \sigma^{2}_{\mathrm{AIPW}}(\bpol, \epol))$, where $\sigma^{2}_{\mathrm{AIPW}}(\pi, \epol)= \mathbb{E}\Big[\sum^{K}_{a=1}\frac{\big(\epol\big(a, X_t\big)\big)^2\nu^*\big(a, X_t\big)}{\bpol(k\mid X_t)}+ \Big(\sum^{K}_{k=1}\epol(a, X_t)f^*(a, X_t) - R(\epol)\Big)^2\Big]$. Next, we consider the case where the behavior probability is time-dependent. When samples have dependency under a behavior probability $\bpol_t(a\mid x, \mathcal{S}_{t-1})$, one of useful OPE estimators is A2IPW estimator defined as $\widehat{R}^{\mathrm{A2IPW}}_T(\epol) = \frac{1}{T}\sum^T_{t=1}\sum^K_{a=1}\epol(a\mid X_t)\Gamma^{\mathrm{AIPW}}_t(a; \bpol_t, \hat{f}_{t-1})$, where $\hat{f}_{t-1}$ is a step-wise consistent estimator of $f^*$ estimated only using samples $\mathcal{S}_{t-1}$. Then, for a converging behavior probability $\bpol_t$ such that $\bpol_t(a\mid x, \mathcal{S}_{t-1}) \xrightarrow{\mathrm{p}} \alpha(a\mid x)$ as $t\to \infty$, \citet{Kato2020} showed that, under some regularity conditions, $\sqrt{T}\left(\widehat{R}^{\mathrm{A2IPW}}_T(\epol)-R(\epol)\right)\xrightarrow{d}\mathcal{N}\left(0, \sigma^{2}_{\mathrm{A2IPW}}(\pi, \epol)\right)$, where $\sigma^{2}_{\mathrm{A2IPW}}(\pi, \epol)=\mathbb{E}\Big[\sum^{K}_{a=1}\frac{\big(\epol\big(a, X_t\big)\big)^2\nu^*\big(a, X_t\big)}{\alpha(k\mid X_t)}+ \Big(\sum^{K}_{k=1}\epol(a, X_t)f^*(a, X_t) - R(\epol)\Big)^2\Big]$. The detailed explanations are in Appendix~\ref{appdx:asymp}. As explained above, under certain conditions, it is known that the AIPW type estimators achieve the efficiency bound (a.k.a semiparametric lower bound), which is the lower bound of the asymptotic MSE of OPE, among regular $\sqrt{T}$-consistent estimators \citep[Theorem 25.20]{VaartA.W.vander1998As}. Let us note that for the best regret order bandit algorithms, we cannot use them as the behavior policy of OPE because the asymptotic variance including $1/\bpol_t$ diverges. We discuss in Appendix~\ref{appdx:det_discuss}.

\section{Categorization based on Evaluation Policy}
\label{sec:cat_eval}
\begin{figure}[h]
\begin{center}
 \includegraphics[width=85mm]{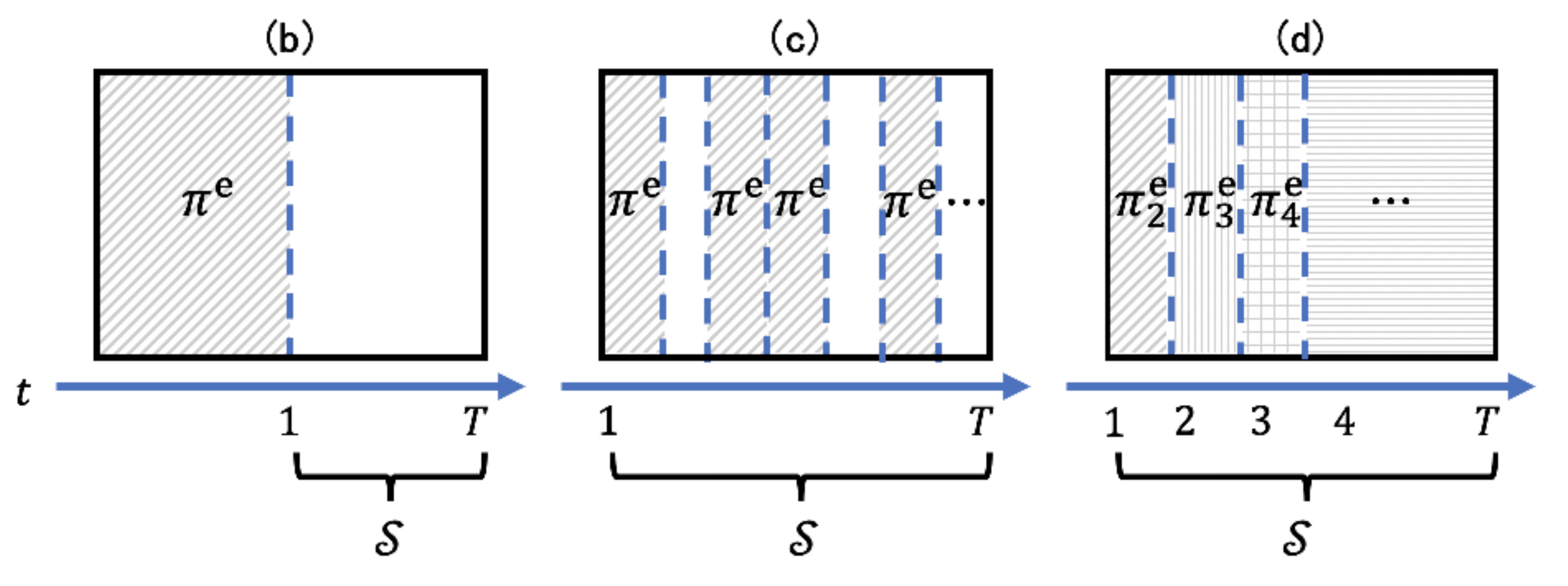}
\end{center}
\caption{Illustrations of cases~(b)--(d).}
\label{fig:concepts}
\end{figure}

In this section, from the deterministic evaluation probability perspective, we categorize the situation where OPE is appropriate as follows: (a) OPE with a given fixed probability; (b) inter-temporal OPE (ITOPE); (c) sample splitting OPE (SSOPE); (d) on-policy off-policy evaluation (OPOPE). While an evaluation probability is given exogenously in case~(a), it is constructed from samples in case~(b)--(d). We illustrate the concepts of case~(b)--(d) in Figure~\ref{fig:concepts}.

When a fixed policy is given as the case~(a), we can naively apply existing OPE methods. However, such a situation is not common in practice because we often construct an evaluation probability from samples. In such cases, we practically consider two cases, (b) ITOPE and (c) SSOPE, where we separate training and evaluation datasets for constructing and evaluating an evaluation probability, respectively. In ITOPE, we have a dataset with time series $\{-T',\dots-1,0,1,\dots, T\}$, where $T'\in\mathbb{N}$, and allow the behavior policy to be updated at period $t=1$ using samples obtained until that period. A defect of these approaches, we cannot use the whole dataset for constructing an evaluation probability. On the other hand, in the case~(d) OPOPE, we can conduct OPE for a sequentially constructed evaluation probability owing to martingale property. In this case, although the whole dataset for constructing an evaluation probability is used for OPE, we only identify the average value of the sequence of policy value. We introduce the details as follows.

\begin{remark}[Categorization based on Behavior Policy]
We also introduce the categorization based on behavior policy from perspectives of time-dependency scenario and the number of behavior policies. The details are shown in Appendix~\ref{appdx:cat_behav_pol}. We show a new theoretical result for the case of multiple behavior policies by using the generalized method of moments (GMM), which enables us to construct an efficient estimator by weighting OPE estimators from the multiple behavior policies. We also show the experimental result using the real-world dataset released by \citet{saito2020large}.
\end{remark}

\paragraph{(a) OPE with a given probability:} In this case, the probability $\epol$ is data-independent and given exogenously. We include ATE estimation in this situation. 

\paragraph{(b) ITOPE:} In this case, we allow behavior and evaluation policies to be time-dependent. In ITOPE, for a time series $t=-T', \dots, -1, 0, 1, \dots, T$, we construct an evaluation probability using past samples obtained until $t=0$. Then, we estimate the policy value using samples from $t=1,2,\dots, T$, $\mathcal{S}$. For representing the dependency, let us denote an evaluation probability as $\epol_{0:-T'}(a\mid x, \mathcal{S}_{0:-T'})$ and behavior probability as $\bpol_{0:-T'}(a\mid x, \mathcal{S}_{0:-T'})$, where $\mathcal{S}_{0:-T'}$  $ =\{X_{0}, A_{0}, Y_{0}, X_{-1}, A_{-1}, Y_{-1}, \ldots, X_{-T'}, A_{-T'}, Y_{-T'}\}$. In this case, we can estimate the policy value defined as $R'_{0:-T'}(\epol) := \mathbb{E}\left[\sum^K_{a=1}\epol(a \mid x, \mathcal{S}_{0:-T'})Y_t(a)\mid \mathcal{S}_{0:-T'}\right]$. This situation is similar to \emph{progressive validation} \citep{blum1999beating}. In this case, an AIPW type estimator $\widehat{R}^{\mathrm{AIPW}'}(\epol_{0:-T'}) = \frac{1}{T}\sum^T_{t=1}\sum^K_{a=1}\epol_{0:-T'}(a\mid X_t, \mathcal{S}_{0:-T'})\Gamma^{\mathrm{AIPW}}_t(a; \bpol_{0:-T'}, \hat{f}_{0:-T'})$ has the asymptotic normality as shown in the following theorem, where $\hat{f}_{0:-T'}$ is an estimator of $f^*$ constructed from $\mathcal{S}_{0:-T'}$.
\begin{theorem}
\label{thm:new_aipw}
Suppose that
\begin{description}
\item[(i)] Pointwise convergence in probability of $\hat{f}_{0:-T'}$, i.e., for all $x\in\mathcal{X}$ and $a\in\{1,2,\dots, K\}$, $\hat{f}_{0:-T'}(a, x)-f^*(a, x)\xrightarrow{\mathrm{p}}0$ as $T'\to\infty$;
\item[(ii)] There exists a constant $C_3$ such that $|\hat{f}_{0:-T'}| \leq C_3$.
\end{description}
Then, under Assumption~\ref{asm:overlap_pol} and \ref{asm:overlap_reward}, 
\begin{align*}
\sqrt{T}\left(\widehat{R}^{\mathrm{AIPW}'}\left(\epol_{0:-T'}\right) - R'(\epol_{0:-T'})\right) \to \mathcal{N}\left(0, \sigma^{2	\dagger}_{\mathrm{AIPW}}\right),
\end{align*}
where $\sigma^{2\dagger}_{\mathrm{AIPW}}$ is the variance of an AIPW estimator with an evaluation probability $\epol_{0:-T}$.
\end{theorem}
The proof is in Appendix~\ref{appdix:new_aipw}. The pointwise convergence assumption of $\hat{f}_{0:-T'}$ is for technical ease of proof.

\paragraph{(c) SSOPE:}
In SSOPE, we randomly separate the dataset $\mathcal{S}$ into multiple subsets. We consider estimating an evaluation policy using a subset and estimate the value using the other subsets. If samples are i.i.d., the estimated evaluation policy is also independent of the evaluation datasets. 

\paragraph{(d) OPOPE:} In OPOPE, we construct an evaluation probability using past samples obtained until period $t-1$ to evaluate the value at each period $t$. In this case, we can estimate the average policy values. Let us denote the evaluation policies as $\epol_t(a\mid x, \mathcal{S}_{t-1})$, and assume that $\pi^{\mathrm{\es}}_t\xrightarrow{\mathrm{p}} \epol$, where $\epol$ is a time invariant probability. Let us define an Average Policy Value AIPW (AP) estimator as $\widehat{R}^{\mathrm{AP}}\big(\{\epol_t\}^T_{t=1}\big) = \frac{1}{T}\sum^T_{t=1}\sum^K_{a=1}\epol_t(a\mid X_t, \mathcal{S}_{t-1})\Gamma^{\mathrm{AIPW}}_t(a; \bpol, \hat{f}_{t-1})$, where note that $\hat{f}_t$ is an estimator of $f^*$ constructed from $\mathcal{S}_{t-1}$. Then, by using the AIPW estimator, we can show the following theorem.

\begin{theorem}
\label{thm:averagea2ipw}
Suppose that
\begin{description}
\item[(i)] Pointwise convergence in probability of $\hat{f}_{t-1}$ and $\epol_t$, i.e., for all $x\in\mathcal{X}$ and $a\in\{1,2,\dots, K\}$, $\hat{f}_{t-1}(a, x)-f^*(a, x)\xrightarrow{\mathrm{p}}0$ and $\epol_t(a\mid x, \mathcal{S}_{t-1})-\epol(a\mid x)\xrightarrow{\mathrm{p}}0$, where $\epol: \mathcal{A}\times\mathcal{X}\to(0,1)$;
\item[(ii)] There exists a constant $C_3$ such that $|\hat{f}_{t-1}| \leq C_3$.
\end{description}
Then, under Assumption~\ref{asm:overlap_pol} and \ref{asm:overlap_reward}, 
\begin{align*}
\sqrt{T}\left(\widehat{R}^{\mathrm{AP}}\left(\{\epol_t\}^T_{t=1}\right) - \frac{1}{T}\sum^T_{t=1}R(\epol_t)\right) \to \mathcal{N}\left(0, \sigma^2_{\mathrm{AP}}\right),
\end{align*}
where $\sigma^2_{\mathrm{AP}}$ is the variance of an AIPW estimator with an evaluation probability $\epol$.
\end{theorem}
The proof is in Appendix~\ref{appdix:averagea2ipw}. In practice, we face such a situation when estimating an evaluation probability from small samples with gathering more samples via bandit process. However, let us note that we cannot estimate the true bandit policy itself because we cannot reproduce the trajectory generated by the target bandit policy Appendix~\ref{appdx:det_discuss}.

\begin{remark}[Estimation of $R(\epol)$]
Can we estimate $R(\epol)$? To this aim, we need to show $\sqrt{T}\frac{1}{T}\sum^T_{t=1}R(\epol_t) - R(\epol) = \mathrm{o}(1)$. This condition requires $\epol_t$ converges to $\epol$ with the $\mathrm{o}(1/\sqrt{T})$ order. However, standard regulation estimators achieve at least $\mathrm{O}(1/\sqrt{T})$ order.
\end{remark}

\begin{remark}[Inadequateness of off-policy bandit evaluation]
In general, it is inadequate to evaluate bandit policies through OPE. The objective of typical bandit policies is to minimize cumulative regret. The value of the regret reflects the transient cumulative cost incurred to reach the optimal stationary policy. To evaluate such a policy, we need to evaluate the policy that also explicitly depends on the history $\mathcal{S}'_{t-1}$, which is different from the history of the log data $\mathcal{S}_{t-1}$ and improbable to reproduce. \citet{gilotte2018offline} also pointed out this inadequateness in their experiments, but did not pursue this topic further.
\end{remark}

\section{Best Evaluation Policy Selection}
\label{sec:beps}
In this section, we consider a case study of the best evaluation policy selection (BEPS), which is one of the most important applications of OPE. For ease of discussion, we only consider an unbiased, consistent, and asymptotically normal estimator under some conditions with deterministic evaluation probability. For example, AIPW and IPW estimators satisfy the requirements. Let us assume that there is a set of $L$-candidate evaluation policies, $\Lambda = \left\{\lambda^{(1)}, \lambda^{(2)}, \dots, \lambda^{(L)}\right\}$, which construct evaluation probabilities $\Pi^{\es} = \{\pi^{\es,\lambda^{(1)}}_{\mathcal{S}}, \pi^{\es,\lambda^{(2)}}_{\mathcal{S}}, \dots, \pi^{\es,\lambda^{(L)}}_{\mathcal{S}}\}$ from a dataset $\mathcal{S}$. In this section, for a given dataset $\mathcal{S}_T$, our goal is to select the best evaluation policy $\lambda^{*}$ with the highest expected reward, i.e., $\lambda^{*} = \argmax_{\lambda\in \Lambda} R(\pi^{\es,\lambda}_{\mathcal{S}_T})$. For achieving this goal, we propose three approaches: (I) estimating the policy value using samples $\mathcal{S}_T$, which also constructs the evaluation probabilities; (II) estimating the policy value using an evaluation dataset independent of the samples constructing the evaluation probability; (III) evaluation the policy value using cross-validation. We refer to these three approaches as In-Sample OPE (ISOPE), OPE with Evaluation Dataset (OPE2D), and Off-Policy Cross-Validation (OPCV). Among them, only OPE2D enables us to estimate the best $\lambda^{*}$ with keeping unbiasedness, consistency, and asymptotic normality. In ISOPE, we cannot guarantee the unbiasedness and asymptotic normality. In OPCV, instead of $R(\pi^{\es,\lambda}_{\mathcal{S}_T})$, we choose the best evaluation policy based on an unbiased, consistent, and asymptotically normal estimator of $R(\pi^{\es,\lambda}_{\mathcal{S}'_T})$, where $\mathcal{S}'_T$ is a subset of $\mathcal{S}_T$. Finally, we discuss the criteria of the BEPS when multiple OPE estimators are given.

\begin{remark}[Off-Policy Learning (OPL)] As a closely related method of BEPS, we introduce OPL, which learns the optimal policy that maximizes the expected reward. Let us define the optimal policy $\pi^*$ as $\pi^* = \argmax_{\pi\in \Pi} R(\pi)$, where $\Pi$ is a policy class. By applying each of OPE estimators $\widehat{R}(\pi)$, we define an estimator for the optimal policy as $\hat{\pi} = \argmax_{\pi\in \mathcal{H}}\widehat{R}(\pi)$, where $\mathcal{H}$ is a hypothesis set. In this case, if $\mathcal{H}$ includes $\pi^*$, we can regard the convergence point of the probability $\hat{\pi}$ as a deterministic one, $\pi^*$.
\end{remark}

\subsection{(I) ISOPE}
Let us consider constructing an OPE estimator from $\mathcal{S}_T$ for estimating a policy value of $\Pi^{\es}$. In this case, as pointed out in the existing studies, in general, we cannot construct an unbiased and asymptotic normal estimator, but can construct a consistent estimator if $\|\pi^{\es,\lambda}_{\mathcal{S}_T}- \pi^{\es*}\|_2 = \mathrm{o}_p(1)$ for a data-independent evaluation probability $\pi^{\es*}$. Let us consider an AIPW estimator $\widehat{R}^{\mathrm{AIPW}}_T(\epol) = \frac{1}{T}\sum^T_{t=1}\sum^K_{a=1}\epol(a | X_t)\Gamma^{\mathrm{AIPW}}_t(a; \bpol_t, \hat{f}_T)$. Then, $\widehat{R}^{\mathrm{AIPW}}_T(\pi^{\es,\lambda}_{\mathcal{S}_T}) - R(\pi^{\es*})$ is equal to $\frac{1}{T}\sum^T_{t=1}\sum^K_{a=1}\left(\pi^{\es,\lambda}_{\mathcal{S}_T}(a | X_t) - \pi^{\es*}(a | X_t)\right)\Gamma^{\mathrm{AIPW}}_t(a; \bpol_t, \hat{f}_T)$ $+ \frac{1}{T}\sum^T_{t=1}\sum^K_{a=1}\pi^{\es*}(a | X_t)\Gamma^{\mathrm{AIPW}}_t(a; \bpol_t, \hat{f}_T) - R(\pi^{\es*})$.
If $\|\pi^{\es,\lambda}_{\mathcal{S}_T}- \pi^{\es*}\|_2 = \mathrm{o}_p(1)$, we can easily show $\widehat{R}^{\mathrm{AIPW}}_T(\pi^{\es,\lambda}_{\mathcal{S}_T}) \xrightarrow{\mathrm{p}} R(\pi^{\es*})$. However, for guaranteeing the asymptotic normality, it is known that we need to impose stronger restrictions for $\pi^{\es,\lambda}_{\mathcal{S}_T}$ \citep{kallus2020stochasticpolicy}. Although this estimator has an asymptotic bias, which does not disappear with $\sqrt{T}$-order \citep{ChernozhukovVictor2018Dmlf}, this estimator is still useful owing to the consistency. ISOPE does not belong to the previous categorization because the evaluation probability is not deterministic.  However, as a special case of ISOPE, if we use OPOPE of case~(d) in Section~\ref{sec:cat_eval}, we can guarantee the unbiasedness to the average of the policy value, but it requires a complicated procedure.

\subsection{(II) OPE2D}
For estimating the policy values of $\Pi^{\es}$, let us consider constructing an OPE estimator from $\mathcal{S}'$, which is independent of $\mathcal{S}_T$. In this case, we can estimate the value of $R(\pi^{\es,\lambda^{(m)}}_{\mathcal{S}_T})$ without loss of the asymptotic normality. This case requires us exogenous dataset $\mathcal{S}'$ but is most desirable from the theoretical perspective. OPE2D is closely related with the case~(b) and (c) in the categorizations of Section~\ref{sec:cat_eval}.

\begin{remark}
In this case, we can also apply hypothesis testing for selecting an evaluation policy. We introduce hypothesis testing framework in Appendix~\ref{appdx:hyp_test}.
\end{remark}

\begin{table*}[t]
\caption{Experimental results of OPE2D with various OPE estimators using mnist and pendigits datasets. The best and worst methods are highlighted in red and blue, respectively.}
\label{tbl:exp_result1}
\scalebox{0.65}[0.65]{
\begin{tabular}{l|rr|rr|rr|rr|rr|rr|rr|rr}
\toprule
{mnist}$\ \ \ \ \ \ \ $ &  \multicolumn{2}{c|}{IPW} &        \multicolumn{2}{c|}{DM LR} &        \multicolumn{2}{c|}{DM KR} &        \multicolumn{2}{c|}{AIPW} &        \multicolumn{2}{c|}{MEAN} &       \multicolumn{2}{c|}{Minimax} &       \multicolumn{2}{c|}{Mix} &       \multicolumn{2}{c}{Maxmax} \\
\hline
$\alpha$ &        Mean &        SD &        Mean &        SD &        Mean &        SD &        Mean &        SD &        Mean &        SD &       Mean &        SD &       Mean &        SD &       Mean &        SD \\
\hline
0.7 &  {0.00239} &  0.00683 &  0.00665 &  0.03627 &  \textcolor{red}{0.00020} &  0.00080 &  0.00248 &  0.00733 &  0.00093 &  0.00542 &  0.00649 &  0.03575 &  \textcolor{blue}{0.00677} &  0.03567 &  0.00181 &  0.00751 \\
0.4 &  0.00006 &  0.00043 &  \textcolor{blue}{0.00430} &  0.01283 &  \textcolor{red}{0.00005} &  0.00040 &  0.00059 &  0.00314 &  \textcolor{red}{0.00005} &  0.00040 &  0.00116 &  0.00608 &  0.00146 &  0.00641 &  0.00306 &  0.01100 \\
0.0 &  \textcolor{red}{0.00000} &  0.00000 &  \textcolor{blue}{0.00690} &  0.04216 &  \textcolor{red}{0.00000} &  0.00000 &  \textcolor{red}{0.00000} &  0.00000 &  0.00437 &  0.03457 &  {0.00545} &  0.04072 &  0.00540 &  0.04070 &  0.00108 &  0.01076 \\
\bottomrule
\end{tabular}}

\scalebox{0.65}[0.65]{
\begin{tabular}{l|rr|rr|rr|rr|rr|rr|rr|rr}
\toprule
{pendigits}$\ $ &  \multicolumn{2}{c|}{IPW} &        \multicolumn{2}{c|}{DM LR} &        \multicolumn{2}{c|}{DM KR} &        \multicolumn{2}{c|}{AIPW} &        \multicolumn{2}{c|}{MEAN} &       \multicolumn{2}{c|}{Minimax} &       \multicolumn{2}{c|}{Mix} &       \multicolumn{2}{c}{Maxmax} \\
\hline
$\alpha$ &        Mean &        SD &        Mean &        SD &        Mean &        SD &        Mean &        SD &        Mean &        SD &       Mean &        SD &       Mean &        SD &       Mean &        SD \\
\hline
0.7 &  0.00170 &  0.00375 &  \textcolor{blue}{0.00448} &  0.00603 &  0.00080 &  0.00208 &  \textcolor{red}{0.00059} &  0.00164 &  0.00113 &  0.00296 &  0.00230 &  0.00453 &  0.00244 &  0.00405 &  0.00145 &  0.00329 \\
0.4 &  0.00075 &  0.00212 &  \textcolor{blue}{0.00535} &  0.00745 &  0.00082 &  0.00247 &  \textcolor{red}{0.00073} &  0.00203 &  0.00093 &  0.00251 &  0.00460 &  0.00696 &  0.00530 &  0.00678 &  0.00088 &  0.00253 \\
0.0 &  0.00061 &  0.00325 &  0.00073 &  0.00368 &  \textcolor{red}{0.00024} &  0.00232 &  \textcolor{red}{0.00024} &  0.00232 &  0.00057 &  0.00337 &  0.00073 &  0.00368 &  \textcolor{blue}{0.00084} &  0.00384 &  0.00054 &  0.00318 \\
\bottomrule
\end{tabular}}
\end{table*}

\subsection{(III) OPCV}
Cross-validation (CV) is a standard method for measuring the performance of methods. For simplicity, the dataset $\mathcal{S}_T$ is randomly separated into two datasets $\mathcal{S}^{(1)}_{T^{(1)}}$ and $\mathcal{S}^{(2)}_{T^{(2)}}$, where note that sample sizes are $T^{(1)}$ and $T^{(2)}$, respectively. Then, using the two datasets, we construct evaluation probabilities using $\mathcal{S}^{(1)}_{T^{(1)}}$ and estimate the policy value by using $\mathcal{S}^{(2)}_{T^{(2)}}$. Next, we construct evaluation probabilities using $\mathcal{S}^{(2)}_{T^{(2)}}$ and estimate the policy value by using $\mathcal{S}^{(1)}_{T^{(1)}}$. Finally, we calculate an average of the policy value estimators and choose an evaluation probability based on the averaged OPE estimator. In this case, we can estimate the policy values of $R(\pi^{\es,\lambda^{(m)}}_{\mathcal{S}^{(1)}_{T^{(1)}}})$ and $R(\pi^{\es,\lambda^{(m)}}_{\mathcal{S}^{(2)}_{T^{(2)}}})$ with asymptotic normality for $m\in\{1,2,\dots, M\}$ although we do estimate the policy value of $R(\pi^{\es,\lambda^{(m)}}_{\mathcal{S}_{T}})$. OPCV belongs to the case~(c) in the categorizations of Section~\ref{sec:cat_eval}.

\begin{remark}[Evaluation without an exogenous dataset]
Note that, in both ISOPE and OPCV, we cannot estimate the policy value $\pi^{\es,\lambda}_{\mathcal{S}_T}$ for an algorithm $\lambda$ with keeping desirable theoretical properties. Although it is difficult to prepare an exogenous dataset as OPE2D, ISOPE and OPCV are realistic alternative approaches in practice.  
\end{remark}

\subsection{Policy Selection Criteria for Multiple OPE Estimators}
Let us assume that there is a set of $E$ OPE estimators, $\mathcal{E}=\left\{\widehat{R}^{(1)}, \dots, \widehat{R}^{(E)}\right\}$. Unlike the standard CV, in OPE, we cannot directly obtain the target metrics of the algorithm, $\frac{1}{T}\sum^{T}_{t=1}\sum^K_{a=1}Y_i(a, X_t)$. Thus, we need to use various OPE estimators approximating $\mathbb{E}\big[\sum^K_{a=1}\epol(a, X_t)Y_t(a)\big]$. Then, which estimator should we use in the case with several OPE estimators? We raise the following three criteria of OPE estimators for BEPS.

\paragraph{Low asymptotic variance estimator:} First, we consider choosing an OPE estimator with the lowest asymptotic variance. The lower bound of the asymptotic variance is given as semiparametric lower bound (Appendix~\ref{appdx:asymp}), and the asymptotic variances of several estimators, such as an AIPW estimator, are known to achieve this lower bound \citep{narita2019counterfactual}. Therefore, from this perspective, a hopeful OPE estimator is an estimator with the lowest asymptotic variance. However, this approach is not useful if several OPE estimators achieve the lower bound. 

\paragraph{Weighted estimator:} Second, we consider a new OPE estimator $\sum^E_{e}w_e\widehat{R}^{(E)}$ from $\mathcal{E}$ by weighting them using a weight $w_e$ such that $\sum^E_e w_e = 1$. A candidate for this weight is $w_e = 1/E$; that is, constructing a new estimator as the average of $\mathcal{E}$. Another candidate is using $e$-th element of $(I^\top_E\Sigma I_E)^{-1}I^\top_E\Sigma$ as the weight $w_e$if they asymptotically follow the multivariate normal distribution with covariance matrix $\Sigma$ \citep{GVK126800421}.
 
\paragraph{Minimax criterion:} The final criterion is a minimax approach. In this criterion, for a dataset $\mathcal{S}$, we select the best evaluation policy as 
\begin{align}
\label{eq:minimax}
\lambda^{*} = \argmin_{\lambda\in\Lambda} \max_{\widehat{R}\in\mathcal{E}} - \widehat{R}\left(\pi^{\es,\lambda}_{\mathcal{S}}\right).
\end{align}

In addition, we consider the following two-player zero-sum game; the player 1 decides the evaluation policy given as a linear combination of $\pi^{\mathrm{e}, \lambda^{(1)}}_{\mathcal{S}}, \dots, \pi^{\mathrm{e}, \lambda^{(L)}}_{\mathcal{S}}$; the player 2 constructs the estimator given as a linear combination of $\hat{R}^{(1)},\dots,\hat{R}^{(E)}$;
The goal of player1/player2 is to maximize/minimize the estimated expected reward of the evaluation policy. Formally, the minmax strategy of player 1 is
\begin{align}
\label{eq:mix}
  p^{*} = \argmax_{p \in \Delta^{L}} \min_{w \in \Delta^{E}} \sum^E_{j=1} -w_j\hat{R}^{(j)}(\pi^{\mathrm{e}}_p),
\end{align}
where $\pi^{\mathrm{e}}_p = \sum^L_{l=1} p_l \pi^{\es,\lambda^{(l)}}_{\mathcal{S}}$ and note that $\Delta^D := \{(p_d) \in [0,1]^K\mid \sum^D_{d=1}p_d=1\}$. Under some mild assumptions, we can compute $p^{*}$ via linear programming:
\begin{theorem}
\label{thm:game}
Suppose that each estimator can be written as $\hat{R}(\pi) = \frac{1}{T} \sum_{t=1}^T\langle \pi(X_t), \Gamma_{t}\rangle$ for $\hat{R}(\pi)\in\mathcal{E}$, where $T$ is the sample size and $\Gamma_t$ is a component of $\hat{R}(\pi)$. Then, $p^{*}$ is the solution of the following problem:
\begin{align*}
  &\max_{z\in\mathbb{R}, p\in\Delta^{L}}\ z\ \ \ \mathrm{s.t.}\ \sum^L_{l=1} p_l \hat{R}(\pi^{\lambda^{(l)}}_{\mathcal{S}}) \geq z,\; \forall \hat{R}\in\mathcal{E}.
\end{align*}
\end{theorem}
Such a minimax criterion is also adapted in existing machine learning studies \citep{wang2015}. This criterion is conservative and aims to avoid the worst result.

\begin{table*}[t]
\caption{Experimental results of ISOPE with various OPE estimators using mnist and pendigits datasets. The best and worst methods are highlighted in red and blue, respectively.}
\label{tbl:exp_result2}
\scalebox{0.65}[0.65]{
\begin{tabular}{l|rr|rr|rr|rr|rr|rr|rr|rr}
\toprule
{mnist}$\ \ \ \ \ \ \ $ &  \multicolumn{2}{c|}{IPW} &        \multicolumn{2}{c|}{DM LR} &        \multicolumn{2}{c|}{DM KR} &        \multicolumn{2}{c|}{AIPW} &        \multicolumn{2}{c|}{MEAN} &       \multicolumn{2}{c|}{Minimax} &       \multicolumn{2}{c|}{Mix} &       \multicolumn{2}{c}{Maxmax} \\
\hline
$\alpha$ &        Mean &        SD &        Mean &        SD &        Mean &        SD &        Mean &        SD &        Mean &        SD &       Mean &        SD &       Mean &        SD &       Mean &        SD \\
\hline
0.7 &  \textcolor{blue}{0.01519} &  0.01420 &  0.00406 &  0.00812 &  \textcolor{red}{0.00294} &  0.00556 &  0.00745 &  0.01130 &  0.00768 &  0.01143 &  0.00572 &  0.01008 &  0.00554 &  0.00924 &  0.01140 &  0.01304 \\
0.4 &  \textcolor{blue}{0.02942} &  0.02778 &  0.00362 &  0.00763 &  0.00384 &  0.00631 &  0.01048 &  0.01306 &  0.00899 &  0.01056 &  \textcolor{red}{0.00344} &  0.00614 &  0.00383 &  0.00627 &  0.02645 &  0.02717 \\
0.0 &  \textcolor{blue}{0.08167} &  0.08505 &  0.00121 &  0.01213 &  \textcolor{red}{0.00000} &  0.00000 &  0.02718 &  0.04801 &  0.02191 &  0.04385 &  0.00121 &  0.01213 &  0.00121 &  0.01213 &  0.08106 &  0.08541 \\
\bottomrule
\end{tabular}}

\scalebox{0.65}[0.65]{
\begin{tabular}{l|rr|rr|rr|rr|rr|rr|rr|rr}
\toprule
{pendigits}$\ $ &  \multicolumn{2}{c|}{IPW} &        \multicolumn{2}{c|}{DM LR} &        \multicolumn{2}{c|}{DM KR} &        \multicolumn{2}{c|}{AIPW} &        \multicolumn{2}{c|}{MEAN} &       \multicolumn{2}{c|}{Minimax} &       \multicolumn{2}{c|}{Mix} &       \multicolumn{2}{c}{Maxmax} \\
\hline
$\alpha$ &        Mean &        SD &        Mean &        SD &        Mean &        SD &        Mean &        SD &        Mean &        SD &       Mean &        SD &       Mean &        SD &       Mean &        SD \\
\hline
0.7 &  0.00375 &  0.00751 &  0.00380 &  0.00550 &  {0.00317} &  0.00537 &  \textcolor{red}{0.00299} &  0.00507 &  0.00319 &  0.00536 &  0.00413 &  0.00752 &  \textcolor{blue}{0.00440} &  0.00731 &  0.00319 &  0.00536 \\
0.4 &  0.00199 &  0.00568 &  0.00470 &  0.00754 &  \textcolor{red}{0.00129} &  0.00355 &  0.00163 &  0.00451 &  0.00163 &  0.00451 &  0.00439 &  0.00738 &  \textcolor{blue}{0.00489} &  0.00730 &  0.00199 &  0.00568 \\
0.0 &  \textcolor{blue}{0.00079} &  0.00496 &  \textcolor{blue}{0.00079} &  0.00496 &  \textcolor{red}{0.00010} &  0.00066 &  0.00037 &  0.00272 &  0.00037 &  0.00272 &  \textcolor{blue}{0.00079} &  0.00496 &  \textcolor{blue}{0.00079} &  0.00496 &  \textcolor{blue}{0.00079} &  0.00496 \\
\bottomrule
\end{tabular}}
\end{table*}

\begin{table*}[t]
\caption{Experimental results of OPCV with various OPE estimators using mnist and pendigits datasets. The best and worst methods are highlighted in red and blue, respectively.}
\label{tbl:exp_result3}
\scalebox{0.65}[0.65]{
\begin{tabular}{l|rr|rr|rr|rr|rr|rr|rr|rr}
\toprule
{mnist}$\ \ \ \ \ \ \ $ &  \multicolumn{2}{c|}{IPW} &        \multicolumn{2}{c|}{DM LR} &        \multicolumn{2}{c|}{DM KR} &        \multicolumn{2}{c|}{AIPW} &        \multicolumn{2}{c|}{MEAN} &       \multicolumn{2}{c|}{Minimax} &       \multicolumn{2}{c|}{Mix} &       \multicolumn{2}{c}{Maxmax} \\
\hline
$\alpha$ &        Mean &        SD &        Mean &        SD &        Mean &        SD &        Mean &        SD &        Mean &        SD &       Mean &        SD &       Mean &        SD &       Mean &        SD \\
\hline
0.7 &  \textcolor{red}{0.02055} &  0.02191 &  0.02318 &  0.02461 &  \textcolor{blue}{0.02386} &  0.02410 &  0.02296 &  0.02148 &  0.02162 &  0.02272 &  0.02197 &  0.02147 &  0.02281 &  0.02097 &  0.02204 &  0.02426 \\
0.4 &  \textcolor{blue}{0.01728} &  0.02157 &  0.01701 &  0.02108 &  0.01653 &  0.02091 &  0.01708 &  0.02168 &  0.01599 &  0.02116 &  \textcolor{red}{0.01588} &  0.02099 &  0.01595 &  0.02019 &  0.01726 &  0.02104 \\
0.0 &  \textcolor{blue}{0.05796} &  0.07521 &  0.05330 &  0.07467 &  0.05369 &  0.07449 &  0.05559 &  0.07465 &  0.05531 &  0.07450 &  \textcolor{red}{0.05217} &  0.07364 &  0.05350 &  0.07421 &  0.05736 &  0.07541 \\
\bottomrule
\end{tabular}}

\scalebox{0.65}[0.65]{
\begin{tabular}{l|rr|rr|rr|rr|rr|rr|rr|rr}
\toprule
{pendigits}$\ $ &  \multicolumn{2}{c|}{IPW} &        \multicolumn{2}{c|}{DM LR} &        \multicolumn{2}{c|}{DM KR} &        \multicolumn{2}{c|}{AIPW} &        \multicolumn{2}{c|}{MEAN} &       \multicolumn{2}{c|}{Minimax} &       \multicolumn{2}{c|}{Mix} &       \multicolumn{2}{c}{Maxmax} \\
\hline
$\alpha$ &        Mean &        SD &        Mean &        SD &        Mean &        SD &        Mean &        SD &        Mean &        SD &       Mean &        SD &       Mean &        SD &       Mean &        SD \\
\hline
0.7 &  \textcolor{red}{0.00689} &  0.01216 &  \textcolor{blue}{0.01872} &  0.02768 &  0.00910 &  0.01761 &  0.00946 &  0.01803 &  0.00903 &  0.01759 &  0.01120 &  0.02113 &  0.01103 &  0.02094 &  0.01672 &  0.02655 \\
0.4 &  \textcolor{red}{0.01384} &  0.02227 &  \textcolor{blue}{0.02459} &  0.03017 &  0.01889 &  0.02723 &  0.01395 &  0.02228 &  0.01773 &  0.02610 &  0.02002 &  0.02736 &  0.01965 &  0.02617 &  0.01903 &  0.02705 \\
0.0 &  \textcolor{red}{0.00078} &  0.00397 &  \textcolor{blue}{0.00722} &  0.04067 &  0.00088 &  0.00406 &  \textcolor{red}{0.00078} &  0.00397 &  0.00088 &  0.00406 &  0.00616 &  0.03941 &  0.00681 &  0.03985 &  \textcolor{red}{0.00078} &  0.00397 \\
\bottomrule
\end{tabular}}
\end{table*}

\section{Experiments}
\label{sec:exp}
In this section, we empirically investigate our arguments.

\subsection{Stochastic Policy Bias}
\begin{figure}[h]
\begin{center}
 \includegraphics[width=85mm]{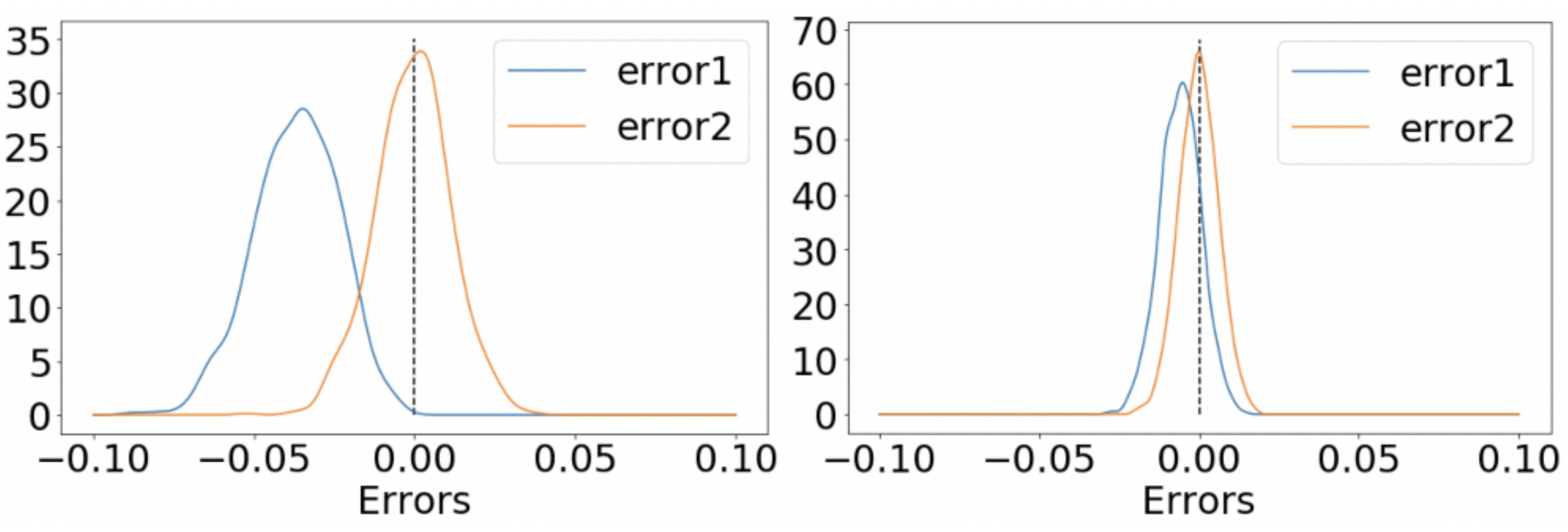}
\end{center}
\caption{The error distributions. We smoothed the histograms using kernel density estimation. The left graph is the results with sample sizes $T^{(1)}=T^{(3)}=100$. The right graph is the results with sample sizes $T^{(1)}=T^{(3)}=1000$.}
\label{fig:bias}
\end{figure}

We investigate what causes by estimating policy value based on a dataset used for constructing evaluation probability.  We generate an artificial pair of covariate and reward $(X_t, Y_t(1), Y_t(2), Y_t(3))$. The covariate $X_t$ is a $10$ dimensional vector generated from the standard normal distribution. For $a\in\{1,2,3\}$, the potential outcome $Y_t(a)$ is $1$ if $a$ is chosen by following a probability defined as $p(a\mid x) = \frac{\exp(g(a, x))}{\sum^3_{a'}\exp(g(a', x))}$, where $g(1, x) = \sum^{10}_{d=1} X_{t,d}$, $g(2, x) = \sum^{10}_{d=1} W_dX^2_{t,d}$, and $g(3, x) = \sum^{10}_{d=1} W_d|X_{t,d}|$, where $W_d$ is uniform randomly chosen from $\{-1, 1\}$. Let us generate three datasets, $\mathcal{S}^{(1)}_{T_{(1)}}$, $\mathcal{S}^{(2)}_{T_{(2)}}$, and $\mathcal{S}^{(3)}_{T_{(3)}}$, where $\mathcal{S}^{(m)}_{T_{(m)}} = \{(X^{(m)}_t, Y^{(m)}_t(1), Y^{(m)}_t(2), Y^{(m)}_t(3))\}^{T^{(m)}}_{t=1}$. Firstly, we train an evaluation probability $\pi^{\es(1)}$ by solving a prediction problem between $X^{(1)}_t$ and $Y^{(1)}_t(1), Y^{(1)}_t(2), Y^{(1)}_t(3)$ using the dataset $\mathcal{S}^{(1)}_{T_{(1)}}$. Then, we apply the evaluation policy $\pi^{\es(1)}$ on the independent dataset $\mathcal{S}^{(2)}_{T_{(2)}}$, and artificially construct bandit data $\{(X^{(m)}_t, A^{(m)}_t, Y^{(m)}_t)\}^{T^{(m)}}_{t=1}$, where $A_t$ is a chosen action from the evaluation policy and $Y^{(m)}_t = \sum^3_{a=1}\mathbbm{1}[A^{(m)}_t = a]Y^{(m)}_t(a)$. Then, we set the true policy value $R(\pi^{\es(1)})$ as $\frac{1}{T^{(2)}}\sum^{T^{(2)}}_{t=1}Y^{(m)}_t$. Next, using the datasets $\mathcal{S}^{(1)}_{T_{(1)}}$ and $\mathcal{S}^{(3)}_{T_{(3)}}$, we estimate the policy value as $\widehat{R}^{(1)}(\pi^{\es(1)}) = \frac{1}{T^{(1)}}\sum^{T^{(1)}}_{t=1}\sum^3_{a=1}\pi^{\es(1)}(a\mid X_t)Y^{(1)}_t(a)$ and $\widehat{R}^{(3)}(\pi^{\es(1)}) = \frac{1}{T^{(3)}}\sum^{T^{(3)}}_{t=1}\sum^3_{a=1}\pi^{\es(1)}(a\mid X_t)Y^{(3)}_t(a)$, respectively. Here, the estimator $\widehat{R}^{(1)}(\pi^{\es(1)})$ violates the assumption of deterministic evaluation policy because $\pi^{\es(1)}$ and $\mathcal{S}^{(1)}_{T_{(1)}}$ are correlated. Let us define estimation errors as $\mathrm{error1} = R(\pi^{\es(1)}) - \widehat{R}^{(1)}(\pi^{\es(1)})$ and $\mathrm{error2} = R(\pi^{\es(1)}) - \widehat{R}^{(3)}(\pi^{\es(1)})$, respectively. We conduct two cases of experiments by changing the sample size. The first case is $T^{(1)} =  T^{(3)} = 100$, and the second case is $T^{(1)}=T^{(3)}=1000$. In both cases, we set $T^{(2)}=10000$. We plot the distributions of these errors in Figure~\ref{fig:bias}. If there is no bias, the error concentrates around on $0$. As expected, we can check that $\mathrm{error1}$ is biased, i.e., $\widehat{R}^{(1)}(\pi^{\es(1)})$ is biased. The results are shown in Figure~\ref{fig:bias}. These results imply that evaluating evaluation probability using samples used for constructing the probability may cause a serious bias even if we observe potential outcomes, and the bias reduces as the sample size increases.

\subsection{OPE2D, ISOPE, and OPCV with Various OPE Estimators}
Following \citet{dudik2011doubly} and \citet{Chow2018}, we evaluate the proposed estimators using classification datasets by transforming them into contextual bandit data. From the LIBSVM repository, we use the pendigits, mnist, sensorless, and connect-4 datasets \footnote{\url{https://www.csie.ntu.edu.tw/~cjlin/libsvmtools/datasets/}}. Then, we make a deterministic policy $\pi_d$ by training a logistic regression classifier on the historical data. We construct three different behavior policies as mixtures of $\pi^d$ and the uniform random policy $\pi^u$ by changing a mixture parameter $\alpha$, i.e., $\bpol = \alpha\pi^d+(1-\alpha)\pi^u$. The candidates of the mixture parameter $\alpha$ are $\{0.7, 0.4, 0.0\}$ as \citet{Kallus2019IntrinsicallyES}. We create a set of $6$ evaluation policies, $\mathcal{E}$, as outputs of linear logistic regression, support vector machine (SVM) with linear kernel, SVM with the polynomial kernel, SVM with RBF kernel, and random forest. In each experiment, from a dataset, we generated sample $1000$ samples for training a behavior policy, $1000$ samples for creating evaluation policies, $1000$ samples for OPE, and $2000$ samples for calculating the true policy value. The more details of the experiment are shown in Appendix~\ref{appdx:exp}. 

For each dataset, we prepare a set of the following policy value estimators: IPW, DM, and AIPW. For the DM estimator, we estimate $f^*$ by linear and kernel ridge regression and denote them DM LR and DM KR, respectively. For the AIPW estimator, we estimate $f^*$ by kernel ridge regression. For selecting an OPE estimator from the set of OPE estimators, we introduce some criteria shown in Section~\ref{sec:beps}. First, we create a new OPE estimator by weighting the OPE estimators with equal weight (MEAN). We also applied minimax criteria of \eqref{eq:minimax} and \eqref{eq:mix} and denote them Minimax and Mix, respectively. Finally, for an evaluation probability class $\Pi$, we select an OPE estimator as $\hat{\epol} = \argmin_{\epol\in\Pi} \min_{\widehat{R}\in\mathcal{E}} - \widehat{R}\left(\epol\right)$ and call it Maxmax. As a performance metric, we use the regret defined as $R(\pi^{\es*}) - R(\hat{\epol})$, where $\pi^{\es'} = \argmax_{\epol\in\mathcal{E}}R(\epol)$ and $\hat{\epol}$ is a chosen evaluation probability by a criterion. We conduct $100$ trials and calculate the means (Mean) and standard deviations (SD) of the regrets. The results are shown in Table~\ref{tbl:exp_result1}. 

We also show experimental results of ISOPE and OPCV with multiple OPE estimators. We show them in Tables~\ref{tbl:exp_result2} and \ref{tbl:exp_result3}, respectively. For ISOPE experiment, unlike OPE2D experiment, we use the same dataset for both constructing and evaluating the evaluation probability. For OPCV experiment, we generated $1000$ samples for training a behavior policy, $2000$ samples for $2$ fold CV, and $2000$ samples for calculating the true policy value. After selecting the best method via CV, we reconstruct an evaluation probability using the whole $2000$ samples and measure its regret. The other settings are identical to the OPE2D experiments.

Note that, only in OPE2D experiments, we can directly evaluate the target policy value. Although the minimax criteria fail to select the best policy in OPE2D experiments, they work well in OPCV experiments. As the IPW estimator achieves both the best and worst performance, it is not easy to choose one OPE estimator. Among the proposed criteria, MEAN works stably in these experiments.

\section{Conclusion}
This study presented a guide for OPE. In OPE, there are various technical problems, but some existing studies tend to ignore them. In this paper, we organized the terminologies of OPE for avoiding the confusion in OPE applications and proposed several methods with new theoretical results. Among the applications, we focus on the BEPS problem and proposed several criteria for the practice of OPE. Finally, we showed empirical investigations of problems related to OPE.

\section*{Ethics Statement}
Although OPE methods are expected to solve various applications, they may also return biased results by the abuse of them. This paper casts concern on this problem and shows a guide for the correct usage. Because OPE methods are potentially extensible to ethically sensitive areas, such as medicine, the OPE methods need to be applied with attention to possible issues that may arise.  

\section*{Acknowledgement}
Kaito Ariu was supported by the Nakajima Foundation Scholarship.

\bibliography{arxiv}
\bibliographystyle{icml2019}

\onecolumn
\appendix

\section{Preliminaries}
\label{appdx:prelim}

\subsection{Mathematical Tools}

\begin{definition}\label{dfn:uniint}[Uniformly Integrable, \citet{GVK126800421}, p.~191]  A sequence $\{A_t\}$ is said to be uniformly integrable if for every $\epsilon > 0$ there exists a number $c>0$ such that 
\begin{align*}
\mathbb{E}[|A_t|\cdot I[|A_t \geq c|]] < \epsilon
\end{align*}
for all $t$.
\end{definition}

\begin{proposition}\label{prp:suff_uniint}[Sufficient Conditions for Uniformly Integrable, \citet{GVK126800421}, Proposition~7.7, p.~191]  (a) Suppose there exist $r>1$ and $M<\infty$ such that $\mathbb{E}[|A_t|^r]<M$ for all $t$. Then $\{A_t\}$ is uniformly integrable. (b) Suppose there exist $r>1$ and $M < \infty$ such that $\mathbb{E}[|b_t|^r]<M$ for all $t$. If $A_t = \sum^\infty_{j=-\infty}h_jb_{t-j}$ with $\sum^\infty_{j=-\infty}|h_j|<\infty$, then $\{A_t\}$ is uniformly integrable.
\end{proposition}

\begin{proposition}[$L^r$ Convergence Theorem, \citet{loeve1977probability}]
\label{prp:lr_conv_theorem}
Let $0<r<\infty$, suppose that $\mathbb{E}\big[|a_n|^r\big] < \infty$ for all $n$ and that $a_n \xrightarrow{\mathrm{p}}a$ as $n\to \infty$. The following are equivalent: 
\begin{description}
\item{(i)} $a_n\to a$ in $L^r$ as $n\to\infty$;
\item{(ii)} $\mathbb{E}\big[|a_n|^r\big]\to \mathbb{E}\big[|a|^r\big] < \infty$ as $n\to\infty$; 
\item{(iii)} $\big\{|a_n|^r, n\geq 1\big\}$ is uniformly integrable.
\end{description}
\end{proposition}

\subsection{Martingale Limit Theorems}
\begin{proposition}
\label{prp:mrtgl_WLLN}[Weak Law of Large Numbers for Martingale, \citet{hall2014martingale}]
Let $\{S_n = \sum^{n}_{i=1} X_i, \mathcal{H}_{t}, t\geq 1\}$ be a martingale and $\{b_n\}$ a sequence of positive constants with $b_n\to\infty$ as $n\to\infty$. Then, writing $X_{ni} = X_i\mathbbm{1}[|X_i|\leq b_n]$, $1\leq i \leq n$, we have that $b^{-1}_n S_n \xrightarrow{\mathrm{p}} 0$ as $n\to \infty$ if 
\begin{description}
\item[(i)] $\sum^n_{i=1}P(|X_i| > b_n)\to 0$;
\item[(ii)] $b^{-1}_n\sum^n_{i=1}\mathbb{E}[X_{ni}\mid \mathcal{H}_{t-1}] \xrightarrow{\mathrm{p}} 0$, and;
\item[(iii)] $b^{-2}_n \sum^n_{i=1}\big\{\mathbb{E}[X^2_{ni}] - \mathbb{E}\big[\mathbb{E}\big[X_{ni}\mid \mathcal{H}_{t-1}\big]\big]^2\big\}\to 0$.
\end{description}
\end{proposition}
\begin{remark} The weak law of large numbers for martingale holds when the random variable is bounded by a constant.
\end{remark}

\begin{proposition}
\label{prp:marclt}[Central Limit Theorem for a Martingale Difference Sequence, \citet{GVK126800421}, Proposition~7.9, p.~194] Let $\{X_t\}^\infty_{t=1}$ be an $n$-dimensional vector martingale difference sequence with $\overline{X}_T=\frac{1}{T}\sum^T_{t=1}X_t$. Suppose that 
\begin{description}
\item[(a)] $\mathbb{E}[X^2_t] = \sigma^2_t$, a positive value with $(1/T)\sum^T_{t=1}\sigma^2_t\to\sigma^2$, a positive value; 
\item[(b)] $\mathbb{E}[|X_t|^r] < \infty$ for some $r>2$;
\item[(c)] $(1/T)\sum^{T}_{t=1}X^2_t\xrightarrow{p}\sigma^2$. 
\end{description}
Then $\sqrt{T} \;\overline{X}_T\xrightarrow{d}\mathcal{N}(\bm{0}, \sigma^2)$.
\end{proposition}

\section{OPE Terminologies}
\label{appdx:term}
In this section, we introduce OPE terminologies used in this paper.

\paragraph{Behavior probability and behavior policy:} First, we distinguish the behavior probability and behavior policy. We call a probability choosing an action, $p_t$, a \emph{behavior probability} and a system generating the behavior probability the \emph{behavior policy}. For example, the contextual bandit algorithm is a behavior policy, which returns a behavior probability based on a trajectory. By distinguishing them, we can clarify the goal of OPE; that is, the goal is to estimate the expected reward given a behavior probability.

\paragraph{Evaluation probability and evaluation policy:} As well as the behavior policy and behavior probability, at an evaluation step, we call a probability choosing an action an \emph{evaluation probability} and a system generating the evaluation probability the \emph{evaluation policy}. 

\paragraph{True bandit policy and pseudo bandit policy evaluations:} Because an MAB algorithm controls a trajectory for balancing exploration and exploitation trade-off, an exact evaluation of a bandit policy requires the generation of the trajectory. Let us define the true bandit policy evaluation as an evaluation of an MAB algorithm with reproducing the trajectory generated from the MAB algorithm. However, in general, it is not easy to produce such a trajectory. Although \citet{narita2019counterfactual} and \citet{saito2020large} attempt to evaluate bandit policies, they fail to conduct the \emph{true bandit policy evaluation}. In their experiments, they run several MAB algorithms on a fixed trajectory. We call such approaches \emph{pseudo bandit policy evaluation} because they do not estimate the expected reward given by the bandit policy, unlike the true bandit policy evaluation. On the other hand, as far as we know, OPE for the true bandit policy evaluation has not been proposed and is an intractable problem. 

\section{Asymptotic Normality of OPE Estimators}
\label{appdx:asymp}
In general, for guaranteeing the asymptotic normality, we need to impose the Donsker's condition on $\hat{f}$. However, by using cross-fitting of DDM, we can show the asymptotic normality only with the convergence rate condition of an estimator of $f^*$ \citep{ChernozhukovVictor2018Dmlf}. In $2$-fold DDM, we first separate the dataset into two subsets. Then, we construct an estimator $\hat{f}$ of $f^*$ from one of the datasets and OPE estimator from the other dataset using the estimator $\hat{f}$. When using the AIPW type estimator, we can show the asymptotic normality only by imposing convergence rate conditions on $\hat{f}$. 

Next, we consider the case where the behavior probability is time-dependent. When samples have dependency under a behavior probability $\bpol_t(a\mid x, \mathcal{S}_{t-1})$, one of useful OPE estimators is Adaptive AIPW  (A2IPW) estimator defined as $\widehat{R}^{\mathrm{A2IPW}}_T(\epol) = \frac{1}{T}\sum^T_{t=1}\sum^K_{a=1}\epol(a\mid X_t)\Gamma^{\mathrm{AIPW}}_t(a; \bpol_t, \hat{f}_t)$, where $\hat{f}_t$ is a step-wise consistent estimator of $f^*$ estimated only using samples $\mathcal{S}_{t-1}$. Then, for a converging behavior probability $\bpol_t$ such that $\bpol_t(a\mid x, \mathcal{S}_{t-1}) \xrightarrow{\mathrm{p}} \alpha(a\mid x)$ as $t\to \infty$, \citet{Kato2020} showed the following proposition.
\begin{proposition}[Asymptotic normality of A2IPW estimator, \citep{Kato2020}]
Suppose that
\begin{description}
\item[(i)] Pointwise convergence in probability of $\hat{f}_{t-1}$ and $\pi_t$, i.e., for all $x\in\mathcal{X}$ and $k\in\mathbb{N}$, $\hat{f}_{t-1}(k, x)-f^*(a, x)\xrightarrow{\mathrm{p}}0$ and $\pi_t(a\mid x, \mathcal{S}_{t-1})-\alpha(a\mid x)\xrightarrow{\mathrm{p}}0$, where $\alpha: \mathcal{A}\times\mathcal{X}\to(0,1)$;
\item[(ii)] There exists a constant $C_3$ such that $|\hat{f}_{t-1}| \leq C_3$.
\end{description}

Then, under Assumptions~\ref{asm:overlap_pol} and \ref{asm:overlap_reward}, we have $\sqrt{T}\left(\widehat{R}^{\mathrm{A2IPW}}_T(\epol)-R(\epol)\right)\xrightarrow{d}\mathcal{N}\left(0, \sigma^2\right)$, where 
\begin{align}
\sigma^{2}_{\mathrm{A2IPW}}(\pi, \epol)  = \mathbb{E}\left[\sum^{1}_{k=0}\frac{\nu^*\big(k, X_t\big)}{\alpha(k\mid X_t)} + \Big(f^*(1, X_t) -  f^*(0, X_t) - R(\epol)\Big)^2\right].
\end{align}
\end{proposition}
Let us note that the consistency of $f_t$ requires different theoretical analysis from the standard case because the samples are not i.i.d. For some specific bandit process, existing studies show consistencies of parametric and nonparametirc estimators \citep{yang2002}. 

\begin{remark}[Semiparametric Lower Bound]
\label{rem:semi_low}
The lower bound of the variance is defined for an estimator of an parameter of interest under some posited models of the DGP. If this posited model is a parametric model, it is equal to \Cramer-Rao lower bound. When this posited model is non or semiparametric model, we can still define a corresponding \Cramer-Rao lower bound \citet{bickel98}. The semiparametric lower bound of the DGP defined in (\ref{eq:DGP}) with $\bpol_1(a\mid x, \mathcal{S}_{0})=\bpol_2(a\mid x, \mathcal{S}_{1})=\cdots=\bpol_T(a\mid x, \mathcal{S}_{T-1})=\bpol(a\mid x)$ is given as follows \citep{narita2019counterfactual}:
\begin{align*}
&\sigma^{2}_{\mathrm{OPT}}(\bpol, \epol) = \mathbb{E}\Bigg[\sum^{K}_{a=1}\frac{\big(\epol(a\mid X)\big)^2\nu^*(a, X_t)}{\bpol(a\mid X_t)} + \left(\sum^{K}_{a=1}\epol(a\mid X)f^*(a, X_t) - \theta_0\right)^2\Bigg].
\end{align*}
\end{remark}

\section{Categorization based on behavior probability}
In this section, we introduce a categorization based on behavior probability.

\label{appdx:cat_behav_pol}
\subsection{Categorization based on the Scenarios of Behavior Policy Update}
For time-dependency of behavior policy generating a behavior probability $\bpol_t(a\mid x)$, we consider the following three cases. The first is the case where the behavior probability is time-invariant, i.e., $\bpol_1(a\mid x, \mathcal{S}_{0})=\bpol_2(a\mid x, \mathcal{S}_{1})=\cdots=\bpol_T(a\mid x, \mathcal{S}_{T-1})=\bpol(a\mid x)$ \citep{Li2011}. In the second and third cases, we consider a behavior policy updates the behavior probabilities at period $t$ using past information $\mathcal{S}_{t-1}$. Following \citet{kato2020confinterval}, we classify the such behavior policies into two patterns, \emph{sequential update policy} and \emph{batch update policy}. Let us assume that a behavior probability only depends on the past information $\mathcal{S}_{t-1}$, and let $\bpol:\mathcal{A}\times\mathcal{X}\times\mathcal{M}_{t-1}\to(0,1)$ be a time-dependent behavior probability. In sequential update policy, the probability $\bpol(a\mid x, \mathcal{S}_{t-1})$ is updated at each period \citep{Laan2008TheCA,Kato2020}. In batch update policy, after using a fixed probability $\bpol(a\mid x, \mathcal{S}_{t-1})$ for some periods without updating, the probability $\bpol(a\mid x, \mathcal{S}_{t-1})$ is updated \citep{Hahn2011,narita2019counterfactual}. Although the sequential update is standard in MAB problems, we often apply batch update in industrial applications such as ad-optimization \citep{perchet2016,narita2019counterfactual}. In this paper, without loss of generality, when discussing the case where a behavior probability is time-dependent, we only consider sequential update policy.

\subsection{Categorization based on the Number of Behavior Probabilities}
\begin{figure}[h]
\begin{center}
 \includegraphics[width=85mm]{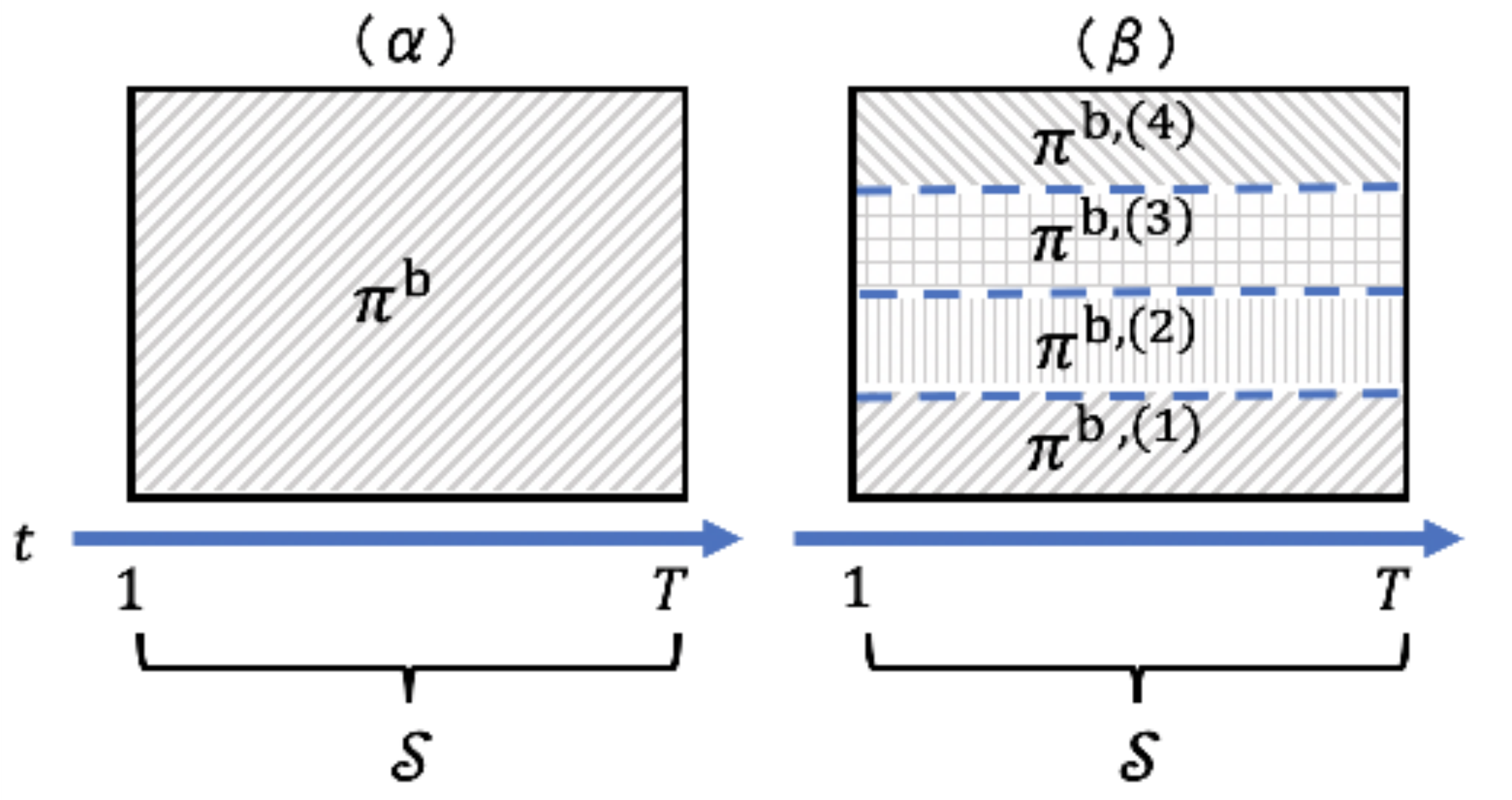}
\end{center}
\caption{Illustrations of cases~($\alpha$) and ($\beta$). In ($\beta$) Multiple behavior probabilities, we assume that there are multiple behavior algorithm on a time-series.}
\label{fig:concepts2}
\end{figure}

The next categorization is based on the number of behavior probabilities. For simplicity, let us consider the case where the behavior probabilities are time-invariant. We can extend the result to the case where the behavior probabilities are time-dependent without loss of generality. The concepts are illustrated in Figure~\ref{fig:concepts2}.

\paragraph{($\alpha$) One behavior probability:} When there is one behavior probability, we use the standard OPE estimator.

\paragraph{($\beta$) Multiple behavior probabilities:} Let us separate the dataset into $M$ subsets, $\big\{\mathcal{S}^{(m)}_{T^{(m)}}\big\}^M_{m=1}$. For each group $m$, we run a behavior probability $\pi^{\bs, (m)}$ as 
\begin{align*}
\mathcal{S}^{(m)}_{T_{(m)}} &= \left\{\left(X^{(m)}_{t}, A^{(m)}_{t}, Y^{(m)}_{t}\right)\right\}^{T^{(m)}}_{{t^{(m)}}=1}\iid p(x)\pi^{\bs, (m)}(a\mid x)p(y\mid a, x),
\end{align*}
where $T^{(m)}$ is the sample size of the dataset $\mathcal{S}^{(m)}_{T^{(m)}}$. When there are $M$ behavior probabilities $\{\pi^{\bs, (m)}\}^M_{m=1}$ for a time series, we can consider two approaches based on the existing OPE estimators.

First, if a sample is uniform randomly grouped into one of $M$ groups, we can define the behavior probability as $\pi^{\bs, \mathrm{mix}} = \frac{1}{M}\sum^M_{m=1}\pi^{\bs, (m)}$. Then, we apply the standard OPE estimators using the created behavior probability $\pi^{\bs, \mathrm{mix}}$. If we use an AIPW estimator under regularity conditions, the asymptotic variance is given as
\begin{align*}
&\sigma^2_{\mathrm{MS}} = \mathbb{E}\Bigg[\sum^{K}_{a=1}\frac{\big(\epol\big(a, X_t\big)\big)^2\nu^*\big(a, X_t\big)}{\pi^{\bs, \mathrm{mix}}(a\mid X_t)} + \Big(\sum^{K}_{a=1}\epol(a, X_t)f^*(a, X_t) - R(\epol)\Big)^2\Bigg],
\end{align*}

Second, we consider using the generalized method of moments (GMM) for $M$ datasets with different behavior probabilities \citet{GVK126800421}. For ease of discussion, we use an AIPW type estimator. Let $D_T$ be an $M$ dimensional vector $(D^{(1)}_T\ \cdots\ D^{(M)}_T)^\top$, where $D^{(m)}_T$ is $\frac{1}{T^{(m)}}\sum^T_{t=1}\sum^K_{a=1}\epol(a\mid X_t)\Gamma^{\mathrm{AIPW}}_t\left(a; \pi^{\bs,(m)}, \hat{f}_T\right)\mathbbm{1}\left[(X_t, A_t, Y_t)\in \mathcal{S}^{(m)}_T\right]$, where $\hat{f}_T$ is a consistent estimator of $f^*$ w.r.t. the sample size $T$. Then, we estimate $R(\epol)$ by $\widehat{R}^{\mathrm{GMM}}(\epol) = \argmin_{R\in\mathbb{R}} (D_T - I_M R)^\top W (D_T - I_MR)$, where $I_M$ is an $M$ dimensional vector $I_M=(1\ 1\ \cdots\ 1)$ and $W$ is an $M\times M$ dimensional positive semidefinite weight matrix. The solution is analytically obtained as $\widehat{R}^{\mathrm{GMM}}_W(\epol) = (I^\top_M W I_M)^{-1} I^\top_M W D_T$. Let us assume that $D^{(m)}_T$ converges to the asymptotic normal distribution with the asymptotic variance $\sigma^2_{(m)}$. Then, for the variance $\sigma^2_{(m)}$, when using a weight function $W^*$ such that $(I^\top_M W^* I_M)^{-1} I^\top_M W^* = \left(\frac{1}{\sigma^2_{(1)}}/\sum^M_{m'=1}\frac{1}{\sigma^2_{(m')}},\dots, \frac{1}{\sigma^2_{(M)}}/\sum^M_{m'=1}\frac{1}{\sigma^2_{(m')}}\right)^\top$, the asymptotic distribution is given as follows.
\begin{theorem}[Asymptotic variance under a stratified sampling]
Suppose that there are $M$ independent datasets, $\big\{\mathcal{S}^{(m)}_T\big\}^M_{m=1}$. If $D_T\xrightarrow{\mathrm{d}}(I_M R(\epol), \Sigma)$, where $\Sigma$ is an $M$ dimensional diagonal matrix with $m$-th diagonal element $\sigma^2_{(m)}$, then $\widehat{R}^{\mathrm{GMM}}_W(\epol)$ asymptotically follows normal distribution with mean $R(\epol)$ and the variance $\sigma^2_{\mathrm{SS}} = \left\{\sum^M_{m=1}\frac{1}{\sigma^2_{(m)}}\right\}^{-1}$,
where 
\begin{align*}
&\sigma^2_{(m)} = \frac{T}{T^{(m)}}\mathbb{E}\Bigg[\sum^{K}_{a=1}\frac{\big(\epol\big(a, X_t\big)\big)^2\nu^*\big(a, X_t\big)}{\pi^{\bs,(m)}(a\mid X_t)} + \Big(\sum^{K}_{a=1}\epol(a, X_t)f^*(a, X_t) - R(\epol)\Big)^2\Bigg].
\end{align*}
\end{theorem}
For example, $D_T\xrightarrow{\mathrm{d}}(I_M R(\epol), \Sigma)$ can be shown when using DDM. Note that, for $m\neq m'$, the covariance is $0$, i.e.,
\begin{align*}
\mathbb{E}\Bigg[&\Gamma^{\mathrm{AIPW}}_t\left(a; \pi^{\bs,(m)}, \hat{f}_T\right)\mathbbm{1}\left[(X_t, A_t, Y_t)\in \mathcal{S}^{(m)}_T\right]\Gamma^{\mathrm{AIPW}}_t\left(a; \pi^{\bs,(m')}, \hat{f}_T\right)\mathbbm{1}\left[(X_t, A_t, Y_t)\in \mathcal{S}^{(m')}_T\right]\Bigg] = 0.
\end{align*}

The difference of the asymptotic variance comes from the difference of DGPs. The first case is called a mixture sampling, and the second case is called a stratified sampling. When considering an estimator $\widehat{R}^{\mathrm{GMM}}(\epol) = \argmin_{R\in\mathbb{R}} (D_T - I_M R)^\top W (D_T - I_MR)$, the first case is a special case of the estimator of the second case with a weight function $W'=(T^{(1)}, \dots, T^{(M)})^\top$. For the estimator such as $\widehat{R}^{\mathrm{GMM}}(\epol)$, when using the weight $W^*$ defined above, the asymptotic variance is minimized \citet{GVK126800421}. Therefore, $\sigma^2_{\mathrm{MS}}\geq  \sigma^2_{\mathrm{SS}}$ holds.

\paragraph{Experiments of GMM:}
\citet{saito2020large} released a dataset for OPE, which is a logged bandit feedback collected on a large-scale fashion e-commerce platform, ZOZOTOWN. The data is generated from Bernoulli Thompson Sampling (BTS) and random policies. The action space is $\mathcal{A}=\{1,2,\dots, 80\}$. The dimension of feature vector is $27$ in BTS policy data and $26$ in random policy data. The sample size of BTS policy data is $12,357,200$ and that of random policy data is $1,374,327$. Let an evaluation policy be $\epol(a\mid X_t) = a/\sum^{80}_{a'=1}a'$. Let us make an evaluation dataset $\big\{\mathcal{S}^{BTS}_{T^{BTS}}, \mathcal{S}^{random}_{T^{random}}\big\}$, where $\mathcal{S}^{BTS}_{T^{BTS}}$ is generated from the BTS policy data, $\mathcal{S}^{random}_{T^{random}}$ is generated from the random policy data, and $T^{BTS}$ and $T^{random}$ denote the sample sizes, respectively. For the dataset, we construct four estimators: AIPW estimator only using $\mathcal{S}^{BTS}_{T^{BTS}}$ (BTSAIPW), AIPW estimator only using $\mathcal{S}^{random}_{T^{random}}\big\}$ (RAIPW), AIPW estimator using mixed behavior policy $\pi^{\mathrm{mix}}$ defined in Section~\ref{sec:cat_eval} (MAIPW), and AIPW based GMM estimator $\widehat{R}^{\mathrm{GMM}}_W(\epol)$ (GMM) defined in Section~4. We calculate the true value of the evaluation policy by using $1,000,000$ data generated from random policy data, which is not used for the evaluation data. For tree dataset with sample sizes $T^{BTS}=T^{random}=100,000$, $T^{BTS}=T^{random}=150,000$, and $T^{BTS}=T^{random}=200,000$, we calculate the root MSEs (RMSEs) between the estimators and the true value and the standard deviations (SD). The result is shown in Table~\ref{tbl:exp_table_mult_pol}. Among the estimators, the GMM estimator achieves the lowest MSEs. 

\begin{table}[h]
\begin{center}
\caption{Experimental results of OPE from multiple behavior policy. The best performing method is highlighted in bold, where * denotes that there is a $95\%$ significant difference compared with the suboptimal method in $t$-test.} 
\medskip
\label{tbl:exp_table_mult_pol}
\scalebox{0.74}[0.74]{
\begin{tabular}{l|rr|rr|rr}
\toprule
sample size &      \multicolumn{2}{c|}{150,000}  &      \multicolumn{2}{c|}{150,000} &   \multicolumn{2}{c}{150,000}\\
\hline
 & RMSE & SD & RMSE & SD & RMSE & SD\\
\hline
BTSAIPW &  0.00151 & 0.00128 &  0.00136 &  0.00105 & 0.00129 & 0.00079\\
RAIPW &  0.00026 & 0.00018 &  0.00026 &  0.00016 & 0.00026 & 0.00015\\
MAIPW &  0.00069 & 0.00064 &  0.00058 &  0.00053 & 0.00055 & 0.00040\\
GMM &  \textbf{0.00021} & 0.00015 &  \textbf{0.00019}* &  0.00014 & \textbf{0.00018} & 0.00013\\
\bottomrule
\end{tabular}
} 
\end{center}
\end{table}

\section{Proof of Theorem~\ref{thm:new_aipw}}
\label{appdix:new_aipw}

For $Z_{t, 0:-T'} = \sum^K_{a=1}\epol_{0:-T'}(a\mid X_t, \mathcal{S}_{t-1})\Gamma^{\mathrm{AIPW}}_t(a; \bpol_{0:-T'}, \hat{f}_{t-1}) - R(\epol_{0:-T'})$, we show
\begin{align*}
\sqrt{T}\left(\frac{1}{T}\sum^T_{t=1}Z_{t, 0:-T'}\right) \to \mathcal{N}\left(0, \sigma^{2\dagger}\right),
\end{align*}

where note that 

\begin{align*}
\Gamma^{\mathrm{AIPW}}_t(a; \bpol, f) = \frac{\mathbbm{1}[A_t = a](Y_t - f(a, X_t))}{\bpol(A_t\mid X_t)} + f(a, X_t).
\end{align*}

Then, the sequence $\{Z_t\}^T_{t=1}$ is an MDS, i.e.,

\begin{align*}
&\mathbb{E}\big[Z_t\mid \mathcal{S}_{t-1}\big]
\\
&= \mathbb{E}\left[\sum^K_{a=1}\epol_{0:-T'}(a\mid X_t, \mathcal{S}_{t-1})\Gamma^{\mathrm{AIPW}}_t(a; \bpol_{0:-T'}, \hat{f}_{t-1}) - R(\epol_t)\mid \mathcal{S}_{t-1}\right]
\\
&= \mathbb{E}\left[\sum^K_{a=1}\epol_{0:-T'}(a\mid X_t, \mathcal{S}_{t-1})\hat{f}_{t-1}(a, X_t) - R(\epol_t)\mid \mathcal{S}_{t-1}\right] + \mathbb{E}\left[\sum^K_{a=1}\frac{\epol_t(a\mid X_t, \mathcal{S}_{t-1})\mathbbm{1}[A_t = a](Y_t - \hat{f}_{t-1}(a, X_t))}{\bpol_{0:-T'}(A_t\mid X_t)}\mid \mathcal{S}_{t-1}\right]\\
&= \mathbb{E}\left[\sum^K_{a=1}\epol_{0:-T'}(a\mid X_t, \mathcal{S}_{t-1})\hat{f}_{t-1}(a, X_t) - R(\epol_t)\mid \mathcal{S}_{0}\right] + \mathbb{E}\left[\sum^K_{a=1}\frac{\epol_t(a\mid X_t, \mathcal{S}_{t-1})\mathbbm{1}[A_t = a](Y_t - \hat{f}_{t-1}(a, X_t))}{\bpol_{0:-T'}(A_t\mid X_t)}\mid \mathcal{S}_{0}\right]\\
&= \mathbb{E}\left[\sum^K_{a=1}\epol_{0:-T'}(a\mid X_t, \mathcal{S}_{t-1})\hat{f}_{t-1}(a, X_t) - R(\epol_t)\mid \mathcal{S}_{0}\right]\\
&\ \ \ + \mathbb{E}\left[\mathbb{E}\left[\sum^K_{a=1}\frac{\epol_{0:-T'}(a \mid X_t, \mathcal{S}_{t-1})\mathbbm{1}[A_t = a](Y_t - \hat{f}_{t-1}(a, X_t))}{\bpol_{0:-T'}(a\mid X_t)}\mid X_t, \mathcal{S}_{0}\right]\mid \mathcal{S}_{0}\right]\\
&= \mathbb{E}\left[\sum^K_{a=1}\epol_{0:-T'}(a\mid X_t, \mathcal{S}_{t-1})\hat{f}_{t-1}(a, X_t) - R(\epol_t) + \sum^K_{a=1}\epol_t(a\mid X_t, \mathcal{S}_{t-1})f^*(a, X_t) - \sum^K_{a=1}\epol_{0:-T'}(a\mid X_t, \mathcal{S}_{t-1})\hat{f}_{t-1}(a, X_t)\mid \mathcal{S}_{0}\right]\\
& = 0.
\end{align*}

Therefore, to derive the asymptotic distribution, we consider applying the CLT for an MDS introduced in Proposition~\ref{prp:marclt}. There are  following three conditions in the statement.

\begin{description}
\item[(a)] $\mathbb{E}\big[Z^2_{t, 0:-T'}\big] = \nu^2_t > 0$ with $\big(1/T\big) \sum^T_{t=1}\nu^2_t\to \nu^2 > 0$;
\item[(b)] $\mathbb{E}\big[|Z_{t, 0:-T'}|^r\big] < \infty$ for some $r>2$;
\item[(c)] $\big(1/T\big)\sum^T_{t=1}Z^2_{t, 0:-T'}\xrightarrow{\mathrm{p}} \nu^2$. 
\end{description}
The proof procedure  is almost same as that of Theorem~\ref{thm:averagea2ipw}. Therefore, we omit showing the above conditions here.

\section{Efficient Experimental Design}
In the case~(b) ITOPE, we have an evaluation probability at a period $t=1$. Therefore, a naive idea for estimating the policy value is to conduct an RCT (A/B testing) using a dataset $\mathcal{S}$. However, as \citet{Laan2008TheCA}, \citet{Hahn2011}, and  \citet{Kato2020} proposed, we can optimize the behavior probability $\bpol$ for minimizing the asymptotic variance of the OPE estimator, which is more efficient than a plain RCT. In this section, following the existing studies of \citet{Laan2008TheCA}, \citet{Hahn2011}, and  \citet{Kato2020}, which propose an experimental design for one evaluation probability, we introduce an efficient experimental design for multiple evaluation probabilities.

Let us consider a situation where there is a set of $M$ evaluation probabilities, $\Pi_t = \left\{\pi^{\mathrm{e}, (1)}, \pi^{\mathrm{e},(2)}, \dots, \pi^{\mathrm{e},(M)}\right\}$ at period $t=1$. Let us assume that we know the variance $\nu^*(a,x)$ before experiment. In Remark~\ref{rem:seq_est}, we introduce a method that weakens this assumption. In this case, our recommendation for designing an efficient experiment via optimizing a behavior probability is to use the behavior probability defined as 

\begin{align}
\label{eq:efficient_pol}
\pi^{\bs*} = \argmin_{\bpol \in \mathcal{H}}\max_{\epol\in \Pi_t}\sigma^{2}_{\mathrm{OPT}}(\bpol, \epol),
\end{align}
where $\mathcal{H}$ is a class of possible behavior probabilities. note that $\sigma^{2}_{\mathrm{OPT}}$ is the semi-parametric lower bound given a behavior probability $\bpol$ and evaluation probability $\epol$. This minimax formulation gives us a behavior probability that minimizes the worst semi-parametric lower bound among $M$-candidate probabilities.

For independent $M$ hypothesis testings on the null hypothesis $R(\pi^{\mathrm{e},(m)}) = 0$ for $m=1,2,\dots, M$, we also show the sample size need for conducting hypothesis testing with power $\beta$ when using AIPW estimator under regularity conditions. 
\begin{corollary}[Sample size for the experiment, \citet{Kato2020}]
Suppose that the effect size $\Delta$ and AIPW estimator has the asymptotic normality with the asymptotic variance achieving the semi-parametric lower bound. Then, the sample size of power $\beta$ test with controlling Type~I error at $\alpha$ is 
\begin{align*}
T_{\beta}(\Delta) = \frac{\max_{\epol\in \Pi_t}\sigma^2_{\mathrm{AIPW}}(\pi^{\bs*}, \epol)}{\Delta^2}\big(z_{1-\alpha/2}-z_{\beta}\big)^2. 
\end{align*}
\end{corollary}

\begin{remark}[Sequential Estimation]
\label{rem:seq_est}
When the variance $\nu^*$ is unknown, we can sequentially estimate the variance and update an estimator of the efficient behavior probability $\pi^{\bs*}(a\mid x)$. For this method in ATE estimation, \citet{Kato2020} showed that an A2IPW estimator has the same asymptotic distribution as AIPW estimator with the known variance.
\end{remark}

\begin{remark}[Parallel RCTs]
We consider the case where we cannot use OPE for estimating the policy value, and need to conduct parallel RCTs for each evaluation probability by separating the dataset without adjusting the behavior probability, i.e., the evaluation probability is equal to the evaluation probability. In this case, we can still optimize the sample size of each RCT for equalizing the power of each test. 
\end{remark}

\section{Proof of Theorem~\ref{thm:averagea2ipw}}
\label{appdix:averagea2ipw}
The procedure of this proof follows \citet{Kato2020}.
\begin{proof}
For $Z_t = \sum^K_{a=1}\epol_t(a\mid X_t, \mathcal{S}_{t-1})\Gamma^{\mathrm{AIPW}}_t(a; \bpol, \hat{f}_{t-1}) - R(\epol_t)$, we show
\begin{align*}
\sqrt{T}\left(\frac{1}{T}\sum^T_{t=1}Z_t\right) \to \mathcal{N}\left(0, \widetilde{\sigma}^2\right),
\end{align*}

where note that 

\begin{align*}
\Gamma^{\mathrm{AIPW}}_t(a; \bpol, f) = \frac{\mathbbm{1}[A_t = a](Y_t - f(a, X_t))}{\bpol(A_t\mid X_t)} + f(a, X_t).
\end{align*}

Then, the sequence $\{Z_t\}^T_{t=1}$ is an MDS, i.e.,

\begin{align*}
&\mathbb{E}\big[Z_t\mid \mathcal{S}_{t-1}\big]
\\
&= \mathbb{E}\left[\sum^K_{a=1}\epol_t(a\mid X_t, \mathcal{S}_{t-1})\Gamma^{\mathrm{AIPW}}_t(a; \pi, \hat{f}_{t-1}) - R(\epol_t)\mid \mathcal{S}_{t-1}\right]
\\
&= \mathbb{E}\left[\sum^K_{a=1}\epol_t(a\mid X_t, \mathcal{S}_{t-1})\hat{f}_{t-1}(a, X_t) - R(\epol_t)\mid \mathcal{S}_{t-1}\right] + \mathbb{E}\left[\sum^K_{a=1}\frac{\epol_t(a\mid X_t, \mathcal{S}_{t-1})\mathbbm{1}[A_t = a](Y_t - \hat{f}_{t-1}(a, X_t))}{\bpol(A_t\mid X_t)}\mid \mathcal{S}_{t-1}\right]
\\
&= \mathbb{E}\left[\sum^K_{a=1}\epol_t(a\mid X_t, \mathcal{S}_{t-1})\hat{f}_{t-1}(a, X_t) - R(\epol_t)\mid \mathcal{S}_{t-1}\right]\\
&\ \ \ + \mathbb{E}\left[\mathbb{E}\left[\sum^K_{a=1}\frac{\epol_t(a \mid X_t, \mathcal{S}_{t-1})\mathbbm{1}[A_t = a](Y_t - \hat{f}_{t-1}(a, X_t))}{\bpol(a\mid X_t)}\mid X_t, \mathcal{S}_{t-1}\right]\mid \mathcal{S}_{t-1}\right]\\
&= \mathbb{E}\left[\sum^K_{a=1}\epol_t(a\mid X_t, \mathcal{S}_{t-1})\hat{f}_{t-1}(a, X_t) - R(\epol_t) + \sum^K_{a=1}\epol_t(a\mid X_t, \mathcal{S}_{t-1})f^*(a, X_t) - \sum^K_{a=1}\epol_t(a\mid X_t, \mathcal{S}_{t-1})\hat{f}_{t-1}(a, X_t)\mid \mathcal{S}_{t-1}\right]\\
& = 0.
\end{align*}

Therefore, to derive the asymptotic distribution, we consider applying the CLT for an MDS introduced in Proposition~\ref{prp:marclt}. There are  following three conditions in the statement.

\begin{description}
\item[(a)] $\mathbb{E}\big[Z^2_t\big] = \nu^2_t > 0$ with $\big(1/T\big) \sum^T_{t=1}\nu^2_t\to \nu^2 > 0$;
\item[(b)] $\mathbb{E}\big[|Z_t|^r\big] < \infty$ for some $r>2$;
\item[(c)] $\big(1/T\big)\sum^T_{t=1}Z^2_t\xrightarrow{\mathrm{p}} \nu^2$. 
\end{description}
Because we assumed the boundedness of $z_t$ by assuming the boundedness of $Y_t$, $\hat{f}_{t-1}$, and $1/\pi_{t}$, the condition~(b) holds. Therefore, the remaining task is to show the conditions~(a) and (c) hold.

\subsection*{Step~1: Check of Condition~(a)}
We can rewrite $\mathbb{E}\big[Z^2_t\big]$ as
\begin{align*}
\mathbb{E}\big[Z^2_t\big]=&\mathbb{E}\left[\left(\sum^K_{a=1}\epol_t(a\mid X_t, \mathcal{S}_{t-1})\left(\frac{\mathbbm{1}[A_t = a](Y_t - f(a, X_t))}{\bpol(a\mid X_t)} + f(a, X_t)\right) - R(\epol_t)\right)^2\right]\\
& -\mathbb{E}\left[\sum^{K}_{a=1}\frac{\big(\epol(a\mid X_t)\big)^2\nu^*\big(a, X_t\big)}{\bpol(a\mid X_t)} + \Big(\sum^{K}_{a=1}\epol(a\mid X_t)f^*(a, X_t) - R(\epol)\Big)^2\right]\\
&+\mathbb{E}\left[\sum^{K}_{a=1}\frac{\big(\epol(a\mid X_t)\big)^2\nu^*\big(a, X_t\big)}{\bpol(a\mid X_t)} + \Big(\sum^{K}_{a=1}\epol(a\mid X_t)f^*(a, X_t) - R(\epol)\Big)^2\right].
\end{align*}
We will prove that the RHS of the following equation varnishes asymptotically to show that the condition~(a) holds.
\begin{align}
\label{eq:1}
&\mathbb{E}\big[Z^2_t\big] - \mathbb{E}\left[\sum^{K}_{a=1}\frac{\big(\epol(a\mid X_t)\big)^2\nu^*\big(a, X_t\big)}{\bpol(a\mid X_t)} + \Big(\sum^{K}_{a=1}\epol(a\mid X_t)f^*(a, X_t) - R(\epol)\Big)^2\right]\nonumber\\
&=\mathbb{E}\left[\left(\sum^K_{a=1}\epol_t(a\mid X_t, \mathcal{S}_{t-1})\left(\frac{\mathbbm{1}[A_t = a](Y_t - \hat{f}_{t-1}(a, X_t))}{\bpol(a\mid X_t)} + \hat{f}_{t-1}(a, X_t)\right) - R(\epol_t)\right)^2\right]\nonumber\\
&\ \ \ -\mathbb{E}\left[\sum^{K}_{a=1}\frac{\big(\epol(a\mid X_t)\big)^2\nu^*\big(a, X_t\big)}{\bpol(a\mid X_t)} + \Big(\sum^{K}_{a=1}\epol(a\mid X_t)f^*(a, X_t) - R(\epol)\Big)^2\right].
\end{align}

First, for the first term of the RHS of \eqref{eq:1},
\begin{align*}
&\mathbb{E}\left[\left(\sum^K_{a=1}\epol_t(a\mid X_t, \mathcal{S}_{t-1})\left(\frac{\mathbbm{1}[A_t = a](Y_t - \hat{f}_{t-1}(a, X_t))}{\bpol(a\mid X_t)} + \hat{f}_{t-1}(a, X_t)\right) - R(\epol_t)\right)^2\right]\\
&=\sum^K_{a=1}\mathbb{E}\left[\left(\frac{\epol_t(a\mid X_t, \mathcal{S}_{t-1})\mathbbm{1}[A_t=a]\big(Y_t - \hat{f}_{t-1}(a, X_t)\big)}{\bpol(a\mid X_t)}\right)^2\right]\\
&\ \ \ +  \mathbb{E}\left[\left(\sum^K_{a=1}\epol_t(a\mid X_t, \mathcal{S}_{t-1})\hat{f}_{t-1}(a, X_t) - R(\epol_t)\right)^2\right]\\
&\ \ \ +2\sum^{K-1}_{a=1}\sum^K_{a'=a+1}\mathbb{E}\left[\left(\frac{\epol_t(a\mid X_t, \mathcal{S}_{t-1})\mathbbm{1}[A_t=a]\big(Y_t - \hat{f}_{t-1}(a, X_t)\big)}{\bpol(a\mid X_t)}\right)\left(\frac{\epol_t(a'\mid X_t, \mathcal{S}_{t-1})\mathbbm{1}[A_t=a']\big(Y_t - \hat{f}_{t-1}(a', X_t)\big)}{\bpol(a'\mid X_t)}\right)\right]\\
&\ \ \ +2\sum^{K}_{a=1}\mathbb{E}\left[\left(\frac{\epol_t(a\mid X_t, \mathcal{S}_{t-1})\mathbbm{1}[A_t=a]\big(Y_t - \hat{f}_{t-1}(a, X_t)\big)}{\bpol(a\mid X_t)}\right)\left(\sum^K_{a'=1}\epol_t(a'\mid X_t, \mathcal{S}_{t-1})\hat{f}_{t-1}(a', X_t) - R(\epol_t)\right)\right].
\end{align*}
Because $\mathbbm{1}[A_t=a]\mathbbm{1}[A_t=a'] = 0$ for $a\neq a'$, $\mathbbm{1}[A_t=a]\mathbbm{1}[A_t=a] = \mathbbm{1}[A_t=a]$, and $\mathbbm{1}[A_t=a]Y_t = Y_t(a)$ for all $a\in\mathcal{A}$ and $a'\neq a$, we have

\begin{align*}
&\mathbb{E}\left[\left(\frac{\epol_t(a\mid X_t, \mathcal{S}_{t-1})\mathbbm{1}[A_t=a]\big(Y_t - \hat{f}_{t-1}(a, X_t)\big)}{\bpol(a\mid X_t)}\right)^2\right]=\mathbb{E}\left[\frac{\left(\epol_t(a\mid X_t, \mathcal{S}_{t-1})\right)^2\big(Y_t(a) - \hat{f}_{t-1}(a, X_t)\big)^2}{\bpol(a\mid X_t)}\right],\\
&\mathbb{E}\left[\left(\frac{\epol_t(a\mid X_t, \mathcal{S}_{t-1})\mathbbm{1}[A_t=a]\big(Y_t - \hat{f}_{t-1}(a, X_t)\big)}{\bpol(a\mid X_t)}\right)\left(\frac{\epol_t(a'\mid X_t, \mathcal{S}_{t-1})\mathbbm{1}[A_t=a']\big(Y_t - \hat{f}_{t-1}(a', X_t)\big)}{\bpol(a'\mid X_t)}\right)\right]=0,\\
&\mathbb{E}\left[\left(\sum^K_{a=1}\epol_t(a\mid X_t, \mathcal{S}_{t-1})\frac{\mathbbm{1}[A_t = a](Y_t - \hat{f}_{t-1}(a, X_t))}{\bpol(a\mid X_t)}\right)\left(\sum^K_{a=1}\epol_t(a\mid X_t, \mathcal{S}_{t-1})\hat{f}_{t-1}(a, X_t) - R(\epol_t)\right)\right]\\
&=\mathbb{E}\left[\mathbb{E}\left[\sum^K_{a=1}\epol_t(a\mid X_t, \mathcal{S}_{t-1})\frac{\mathbbm{1}[A_t = a](Y_t - \hat{f}_{t-1}(a, X_t))}{\bpol(a\mid X_t)}\mid X_t, \mathcal{S}_{t-1}\right]\left(\sum^K_{a=1}\epol_t(a\mid X_t, \mathcal{S}_{t-1})\hat{f}_{t-1}(a, X_t) - R(\epol_t)\right)\right]\\
&=\mathbb{E}\left[\left(\sum^K_{a=1}\epol_t(a\mid X_t, \mathcal{S}_{t-1})f^*(a, X_t) - \sum^K_{a=1}\epol_t(a\mid X_t, \mathcal{S}_{t-1})\hat{f}_{t-1}(a, X_t)\right)\left(\sum^K_{a=1}\epol_t(a\mid X_t, \mathcal{S}_{t-1})\hat{f}_{t-1}(a, X_t) - R(\epol_t)\right)\right].
\end{align*}

Therefore, for the first term of the RHS of \eqref{eq:1}, 

\begin{align*}
&\mathbb{E}\left[\left(\sum^K_{a=1}\epol_t(a\mid X_t, \mathcal{S}_{t-1})\left(\frac{\mathbbm{1}[A_t = a](Y_t - \hat{f}_{t-1}(a, X_t))}{\bpol(a\mid X_t)} + \hat{f}_{t-1}(a, X_t)\right) - R(\epol_t)\right)^2\right]\\
&= \mathbb{E}\Bigg[\sum^K_{a=1}\frac{\left(\epol_t(a\mid X_t, \mathcal{S}_{t-1})\right)^2\big(Y_t(a) - \hat{f}_{t-1}(a, X_t)\big)^2}{\bpol(a\mid X_t)} + \left(\sum^K_{a=1}\epol_t(a\mid X_t, \mathcal{S}_{t-1})\hat{f}_{t-1}(a, X_t) - R(\epol_t)\right)^2+ \\
&\ \ \ 2\left(\sum^K_{a=1}\epol_t(a\mid X_t, \mathcal{S}_{t-1})f^*(a, X_t) - \sum^K_{a=1}\epol_t(a\mid X_t, \mathcal{S}_{t-1})\hat{f}_{t-1}(a, X_t)\right)\left(\sum^K_{a=1}\epol_t(a\mid X_t, \mathcal{S}_{t-1})\hat{f}_{t-1}(a, X_t) - R(\epol_t)\right)\Bigg].
\end{align*}

and, for the second term of the RHS of \eqref{eq:1},

\begin{align*}
&\mathbb{E}\left[\sum^{K}_{a=1}\frac{\big(\epol(a\mid X_t)\big)^2\nu^*\big(a, X_t\big)}{\bpol(a\mid X_t)} + \Big(\sum^{K}_{a=1}\epol(a\mid X_t)f^*(a, X_t) - R(\epol)\Big)^2\right]\\
&=\mathbb{E}\left[\sum^{K}_{a=1}\frac{\big(\epol(a\mid X_t)\big)^2\big(Y_t(a)- f^*(a, X_t)\big)^2}{\bpol(a\mid X_t)} + \Big(\sum^{K}_{a=1}\epol(a\mid X_t)f^*(a, X_t) - R(\epol)\Big)^2\right].
\end{align*}

Then, using these equations, the RHS of \eqref{eq:1} can be calculated as 

\begin{align*}
&\mathbb{E}\left[\left(\sum^K_{a=1}\epol_t(a\mid X_t, \mathcal{S}_{t-1})\left(\frac{\mathbbm{1}[A_t = a](Y_t - \hat{f}_{t-1}(a, X_t))}{\bpol(a\mid X_t)} + \hat{f}_{t-1}(a, X_t)\right) - R(\epol_t)\right)^2\right]\\
&\ \ \ \ \ \ \ \ \ \ \ -\mathbb{E}\left[\sum^{K}_{a=1}\frac{\big(\epol(a\mid X_t)\big)^2\nu^*\big(a, X_t\big)}{\bpol(a\mid X_t)} + \Big(\sum^{K}_{a=1}\epol(a\mid X_t)f^*(a, X_t) - R(\epol)\Big)^2\right]\\
&=\mathbb{E}\Bigg[\sum^K_{a=1}\frac{\left(\epol_t(a\mid X_t, \mathcal{S}_{t-1})\right)^2\big(Y_t(a) - \hat{f}_{t-1}(a, X_t)\big)^2}{\bpol(a\mid X_t)} + \left(\sum^K_{a=1}\epol_t(a\mid X_t, \mathcal{S}_{t-1})\hat{f}_{t-1}(a, X_t) - R(\epol_t)\right)^2+ \\
&\ \ \ \ \ \ \ \ \ 2\left(\sum^K_{a=1}\epol_t(a\mid X_t, \mathcal{S}_{t-1})f^*(a, X_t) - \sum^K_{a=1}\epol_t(a\mid X_t, \mathcal{S}_{t-1})\hat{f}_{t-1}(a, X_t)\right)\left(\sum^K_{a=1}\epol_t(a\mid X_t, \mathcal{S}_{t-1})\hat{f}_{t-1}(a, X_t) - R(\epol_t)\right)\Bigg]\\
&\ \ \ -\mathbb{E}\left[\sum^{K}_{a=1}\frac{\big(\epol(a\mid X_t)\big)^2\big(Y_t(a)- f^*(a, X_t)\big)^2}{\bpol(a\mid X_t)} + \Big(\sum^{K}_{a=1}\epol(a\mid X_t)f^*(a, X_t) - R(\epol)\Big)^2\right].
\end{align*}
Then, from the triangle inequality, we have

\begin{align*}
&\mathbb{E}\Bigg[\sum^K_{a=1}\frac{\left(\epol_t(a\mid X_t, \mathcal{S}_{t-1})\right)^2\big(Y_t(a) - \hat{f}_{t-1}(a, X_t)\big)^2}{\bpol(a\mid X_t)} + \left(\sum^K_{a=1}\epol_t(a\mid X_t, \mathcal{S}_{t-1})\hat{f}_{t-1}(a, X_t) - R(\epol_t)\right)^2+ \\
&\ \ \ \ \ \ \ \ \ 2\left(\sum^K_{a=1}\epol_t(a\mid X_t, \mathcal{S}_{t-1})f^*(a, X_t) - \sum^K_{a=1}\epol_t(a\mid X_t, \mathcal{S}_{t-1})\hat{f}_{t-1}(a, X_t)\right)\left(\sum^K_{a=1}\epol_t(a\mid X_t, \mathcal{S}_{t-1})\hat{f}_{t-1}(a, X_t) - R(\epol_t)\right)\Bigg]\\
&\ \ \ - \mathbb{E}\left[\sum^{K}_{a=1}\frac{\big(\epol(a\mid X_t)\big)^2\big(Y_t(a)- f^*(a, X_t)\big)^2}{\bpol(a\mid X_t)} + \Big(\sum^{K}_{a=1}\epol(a\mid X_t)f^*(a, X_t) - R(\epol)\Big)^2\right]\\
&\leq \sum^K_{a=1}\mathbb{E}\Bigg[\Bigg|\frac{\left(\epol_t(a\mid X_t, \mathcal{S}_{t-1})\right)^2\big(Y_t(a) - \hat{f}_{t-1}(a, X_t)\big)^2}{\bpol(a\mid X_t)} - \frac{\big(\epol(a\mid X_t)\big)^2\big(Y_t(a)- f^*(a, X_t)\big)^2}{\bpol(a\mid X_t)}\Bigg|\Bigg]\\
&\ \ \ + \mathbb{E}\Bigg[\Bigg|\left(\sum^K_{a=1}\epol_t(a\mid X_t, \mathcal{S}_{t-1})\hat{f}_{t-1}(a, X_t) - R(\epol_t)\right)^2 - \Big(\sum^{K}_{a=1}\epol(a\mid X_t)f^*(a, X_t) - R(\epol)\Big)^2\Bigg|\Bigg]\\
&\ \ \ + 2\mathbb{E}\Bigg[\Bigg|\left(\sum^K_{a=1}\epol_t(a\mid X_t, \mathcal{S}_{t-1})f^*(a, X_t) - \sum^K_{a=1}\epol_t(a\mid X_t, \mathcal{S}_{t-1})\hat{f}_{t-1}(a, X_t)\right)\left(\sum^K_{a=1}\epol_t(a\mid X_t, \mathcal{S}_{t-1})\hat{f}_{t-1}(a, X_t) - R(\epol_t)\right)\Bigg|\Bigg].
\end{align*}

Because all elements are assumed to be bounded and $b_1^2 - b_2^2 = (b_1+b_2)(b_1-b_2)$ for variables $b_1$ and $b_2$, there exist constants $\tilde C_0$, $\tilde C_1$, $\tilde C_2$, and $\tilde C_3$ such that

\begin{align*}
& \sum^K_{a=1}\mathbb{E}\Bigg[\Bigg|\frac{\left(\epol_t(a\mid X_t, \mathcal{S}_{t-1})\right)^2\big(Y_t(a) - \hat{f}_{t-1}(a, X_t)\big)^2}{\bpol(a\mid X_t)} - \frac{\epol(a\mid X_t)\big(Y_t(a)- f^*(a, X_t)\big)^2}{\bpol(a\mid X_t)}\Bigg|\Bigg]\\
&\ \ \ + \mathbb{E}\Bigg[\Bigg|\left(\sum^K_{a=1}\epol_t(a\mid X_t, \mathcal{S}_{t-1})\hat{f}_{t-1}(a, X_t) - R(\epol_t)\right)^2 - \Big(\sum^{K}_{a=1}\epol(a\mid X_t)f^*(a, X_t) - R(\epol)\Big)^2\Bigg|\Bigg]\\
&\ \ \ + 2\mathbb{E}\Bigg[\Bigg|\left(\sum^K_{a=1}\epol_t(a\mid X_t, \mathcal{S}_{t-1})f^*(a, X_t) - \sum^K_{a=1}\epol_t(a\mid X_t, \mathcal{S}_{t-1})\hat{f}_{t-1}(a, X_t)\right)\left(\sum^K_{a=1}\epol_t(a\mid X_t, \mathcal{S}_{t-1})\hat{f}_{t-1}(a, X_t) - R(\epol_t)\right)\Bigg|\Bigg]\\
& \leq \tilde C_0\sum^K_{a=1}\mathbb{E}\Bigg[\Bigg|\frac{\epol_t(a\mid X_t, \mathcal{S}_{t-1})\big(Y_t(a) - \hat{f}_{t-1}(a, X_t)\big)}{\sqrt{\bpol(a\mid X_t)}} - \frac{\epol(a\mid X_t)\big(Y_t(a)- f^*(a, X_t)\big)}{\sqrt{\bpol(a\mid X_t)}}\Bigg|\Bigg]\\
&\ \ \ + \mathbb{E}\Bigg[\Bigg|\left(\sum^K_{a=1}\epol_t(a\mid X_t, \mathcal{S}_{t-1})\hat{f}_{t-1}(a, X_t) - R(\epol_t)\right)^2 - \Big(\sum^{K}_{a=1}\epol(a\mid X_t)f^*(a, X_t) - R(\epol)\Big)^2\Bigg|\Bigg]\\
&\ \ \ + 2\mathbb{E}\Bigg[\Bigg|\left(\sum^K_{a=1}\epol_t(a\mid X_t, \mathcal{S}_{t-1})f^*(a, X_t) - \sum^K_{a=1}\epol_t(a\mid X_t, \mathcal{S}_{t-1})\hat{f}_{t-1}(a, X_t)\right)\left(\sum^K_{a=1}\epol_t(a\mid X_t, \mathcal{S}_{t-1})\hat{f}_{t-1}(a, X_t) - R(\epol_t)\right)\Bigg|\Bigg]\\
& \leq \tilde C_1\sum^K_{a=1}\mathbb{E}\Bigg[\Bigg|\epol_t(a\mid X_t, \mathcal{S}_{t-1})\big(Y_t(a) - \hat{f}_{t-1}(a, X_t)\big) - \epol(a\mid X_t)\big(Y_t(a)- f^*(a, X_t)\big)\Bigg|\Bigg]\\
&\ \ \ + \mathbb{E}\Bigg[\Bigg|\left(\sum^K_{a=1}\epol_t(a\mid X_t, \mathcal{S}_{t-1})\hat{f}_{t-1}(a, X_t) - R(\epol_t)\right)^2 - \Big(\sum^{K}_{a=1}\epol(a\mid X_t)f^*(a, X_t) - R(\epol)\Big)^2\Bigg|\Bigg]\\
&\ \ \ + 2\mathbb{E}\Bigg[\Bigg|\left(\sum^K_{a=1}\epol_t(a\mid X_t, \mathcal{S}_{t-1})f^*(a, X_t) - \sum^K_{a=1}\epol_t(a\mid X_t, \mathcal{S}_{t-1})\hat{f}_{t-1}(a, X_t)\right)\left(\sum^K_{a=1}\epol_t(a\mid X_t, \mathcal{S}_{t-1})\hat{f}_{t-1}(a, X_t) - R(\epol_t)\right)\Bigg|\Bigg]\\
&\leq \tilde{C}_1\sum^K_{a=1}\mathbb{E}\left[\Big| \epol_t(a\mid X_t, \mathcal{S}_{t-1})\hat{f}_{t-1}(a, X_t) - \epol(a\mid X_t)f^*(k, X_t) \Big|\right]\\
&\ \ \ + \tilde{C}_2 \sum^K_{a=1}\mathbb{E}\left[\Big| \epol_t(a\mid X_t, \mathcal{S}_{t-1}) - \epol(a\mid X_t) \Big|\right] + \tilde{C}_3 \sum^K_{a=1}\mathbb{E}\left[\Big| \hat{f}_{t-1}(a, X_t) - f^*(a, X_t) \Big|\right].
\end{align*}
Then, from $b_1b_2 - b_3b_4 = (b_1 - b_3)b_4 - (b_4 - b_2)b_1$ for variables $b_1$, $b_2$, $b_3$, and $b_4$, there exist $\tilde C_4$ and $\tilde C_5$ such that 

\begin{align*}
&\tilde{C}_1\sum^K_{a=1}\mathbb{E}\left[\Big| \epol_t(a\mid X_t, \mathcal{S}_{t-1})\hat{f}_{t-1}(a, X_t) - \epol(a\mid X_t)f^*(k, X_t) \Big|\right]\\
&\ \ \ + \tilde{C}_2 \sum^K_{a=1}\mathbb{E}\left[\Big| \epol_t(a\mid X_t, \mathcal{S}_{t-1}) - \epol(a\mid X_t) \Big|\right] + \tilde{C}_3 \sum^K_{a=1}\mathbb{E}\left[\Big| \hat{f}_{t-1}(a, X_t) - f^*(a, X_t) \Big|\right]\\
&\leq \tilde{C}_4 \sum^K_{a=1}\mathbb{E}\left[\Big| \epol_t(a\mid X_t, \mathcal{S}_{t-1}) - \epol(a\mid X_t) \Big|\right] + \tilde{C}_5 \sum^K_{a=1}\mathbb{E}\left[\Big| \hat{f}_{t-1}(a, X_t) - f^*(a, X_t) \Big|\right].
\end{align*}

From the assumption that the pointwise convergences in probability, i.e., for all $x\in\mathcal{X}$ and $k\in\mathcal{A}$, $\epol_t(a\mid X_t, \mathcal{S}_{t-1}) - \epol(a\mid X_t)\xrightarrow{\mathrm{p}}0$ and $\hat{f}_{t-1}(a, x) - f^*(a, x)\xrightarrow{\mathrm{p}}0$ as $t\to\infty$, if $\epol_t(a\mid X_t, \mathcal{S}_{t-1})$, and $\hat{f}_{t-1}(a, x)$ are  uniformly integrable, for fixed $x\in \mathcal{X}$, we can prove that 

\begin{align*}
&\mathbb{E}\big[|\epol_t(a\mid X_t, \mathcal{S}_{t-1}) - \epol(a\mid X_t)| \mid X_t=x\big] = \mathbb{E}\big[|\epol_t(a\mid x, \mathcal{S}_{t-1}) - \epol(a\mid x)|\big]  \to 0,\\
&\mathbb{E}\big[|\hat{f}_{t-1}(a, X_t) - f^*(a, X_t)|\mid X_t = x\big] = \mathbb{E}\big[|\hat{f}_{t-1}(a, x) - f^*(a, x)|\big] \to 0,
\end{align*}
as $t\to\infty$ using $L^r$-convergence theorem (Proposition~\ref{prp:lr_conv_theorem}). Here, we used the fact that $\hat{f}_{t-1}(a, x)$ and $\epol_t(a\mid x, \mathcal{S}_{t-1})$ are independent from $X_t$. For fixed $x\in\mathcal{X}$, we can show that $\epol_t(a\mid x, \mathcal{S}_{t-1})$ and $\hat{f}_{t-1}(a, x)$ are uniformly integrable from the boundedness of $\epol_t(a\mid x, \mathcal{S}_{t-1})$, and $\hat{f}_{t-1}(a, x)$ (Proposition~\ref{prp:suff_uniint}). From the pointwise convergence of $\mathbb{E}[|\epol_t(a\mid X_t, \mathcal{S}_{t-1}) - \epol(a\mid X_t)| \mid X_t=x]$ and $\mathbb{E}[|\hat{f}_{t-1}(a, X_t) - f^*(a, X_t)|\mid X_t = x]$, by using the Lebesgue's dominated convergence theorem, we can show that

\begin{align*}
&\mathbb{E}_{X_t}\big[\mathbb{E}\big[|\epol_t(a\mid X_t, \mathcal{S}_{t-1}) - \epol(a\mid X_t)| \mid X_t\big]\big]  \to 0,\\
&\mathbb{E}_{X_t}\big[\mathbb{E}[|\hat{f}_{t-1}(a, X_t) - f^*(a, X_t)|\mid X_t]\big]  \to 0.
\end{align*}
Then, as $t\to\infty$,

\begin{align*}
&\mathbb{E}\big[Z^2_t\big] - \mathbb{E}\left[\sum^{K}_{a=1}\frac{\big(\epol(a\mid X_t)\big)^2\nu^*\big(a, X_t\big)}{\bpol(a\mid X_t)} + \Big(\sum^{K}_{a=1}\epol(a\mid X_t)f^*(a, X_t) - R(\epol)\Big)^2\right]\to 0.
\end{align*}
Therefore, for any $\epsilon > 0$, there exists $\tilde t > 0$ such that 

\begin{align*}
&\frac{1}{T} \sum^{T}_{t=1}\Bigg(\mathbb{E}\big[Z^2_t\big] - \mathbb{E}\left[\sum^{K}_{a=1}\frac{\big(\epol(a\mid X_t)\big)^2\nu^*\big(a, X_t\big)}{\bpol(a\mid X_t)} + \Big(\sum^{K}_{a=1}\epol(a\mid X_t)f^*(a, X_t) - R(\epol)\Big)^2\right]\Bigg)\leq \frac{\tilde t}{T} + \epsilon.
\end{align*}
Here, 

\begin{align*}
& \mathbb{E}\left[\sum^{K}_{a=1}\frac{\big(\epol(a\mid X_t)\big)^2\nu^*\big(a, X_t\big)}{\bpol(a\mid X_t)} + \Big(\sum^{K}_{a=1}\epol(a\mid X_t)f^*(a, X_t) - R(\epol)\Big)^2\right]
\\
& \ \ \ =  \mathbb{E}\left[\sum^{K}_{a=1}\frac{\big(\epol(a\mid X)\big)^2\nu^*\big(a, X\big)}{\bpol(a\mid X)} + \Big(\sum^{K}_{a=1}\epol(a\mid X)f^*(a, X) - R(\epol)\Big)^2\right]
\end{align*}
does not depend on periods. Therefore, $\big(1/T\big) \sum^{T}_{t=1}\sigma^2_t- \sigma^2 \leq \tilde t/T + \epsilon \to 0$ as $T\to\infty$, where 

\begin{align*}
\sigma^2_{\mathrm{AP}} = \mathbb{E}\left[\sum^{K}_{a=1}\frac{\big(\epol(a\mid X)\big)^2\nu^*\big(a, X\big)}{\bpol(a\mid X)} + \Big(\sum^{K}_{a=1}\epol(a\mid X)f^*(a, X) - R(\epol)\Big)^2\right].
\end{align*}

\subsection*{Step~2: Check of Condition~(c)}
Let $U_t$ be an MDS such that

\begin{align*}
&U_t = Z^2_t - \mathbb{E}\big[Z^2_t\mid \mathcal{S}_{t-1}\big]\\
&=\left(\sum^K_{a=1}\frac{\epol_t(a\mid X_t, \mathcal{S}_{t-1})\mathbbm{1}[A_t = a](Y_t - \hat{f}_{t-1}(a, X_t))}{\bpol(A_t\mid X_t)} + \sum^K_{a=1}\epol_t(a\mid X_t, \mathcal{S}_{t-1})\hat{f}_{t-1}(a, X_t) - R(\epol_t) \right)^2\\\
& - \mathbb{E}\left[\left(\sum^K_{a=1}\frac{\epol_t(a\mid X_t, \mathcal{S}_{t-1})\mathbbm{1}[A_t = a](Y_t - \hat{f}_{t-1}(a, X_t))}{\bpol(A_t\mid X_t)} + \sum^K_{a=1}\epol_t(a\mid X_t, \mathcal{S}_{t-1})\hat{f}_{t-1}(a, X_t) - R(\epol_t)\right)^2\mid \mathcal{S}_{t-1}\right].
\end{align*}
From the boundedness of each variable in $Z_t$, we can apply weak law of large numbers for an MDS (Proposition~\ref{prp:mrtgl_WLLN} in Appendix~\ref{appdx:prelim}), and obtain

\begin{align*}
&\frac{1}{T}\sum^T_{t=1}U_t = \frac{1}{T}\sum^T_{t=1}\big(Z^2_t - \mathbb{E}\big[Z^2_t\mid \mathcal{S}_{t-1}\big]\big)\xrightarrow{\mathrm{p}} 0.
\end{align*}
Next, we show that

\begin{align*}
\frac{1}{T}\sum^T_{t=1}\mathbb{E}\big[Z^2_t\mid \mathcal{S}_{t-1}\big] - \sigma^2_{\mathrm{AP}}\xrightarrow{\mathrm{p}} 0.
\end{align*}
From Markov's inequality, for any $\varepsilon > 0$, we have

\begin{align*}
&\mathbb{P}\left(\left|\frac{1}{T}\sum^T_{t=1}\mathbb{E}\big[Z^2_t\mid \mathcal{S}_{t-1}\big] - \sigma^2_{\mathrm{AP}}\right| \geq \varepsilon\right)\\
&\leq \frac{\mathbb{E}\left[\left|\frac{1}{T}\sum^T_{t=1}\mathbb{E}\big[Z^2_t\mid \mathcal{S}_{t-1}\big] - \sigma^2_{\mathrm{AP}}\right|\right]}{\varepsilon}\\
&\leq \frac{\frac{1}{T}\sum^T_{t=1}\mathbb{E}\left[\big|\mathbb{E}\big[Z^2_t\mid \mathcal{S}_{t-1}\big] - \sigma^2_{\mathrm{AP}}\big|\right]}{\varepsilon}.
\end{align*}
Then, we consider showing $\mathbb{E}\left[\big|\mathbb{E}\big[Z^2_t\mid \mathcal{S}_{t-1}\big] - \sigma^2_{\mathrm{AP}}\big|\right] \to 0$. Here, we have

\begin{align*}
&\mathbb{E}\left[\big|\mathbb{E}\big[Z^2_t\mid \mathcal{S}_{t-1}\big] - \sigma^2_{\mathrm{AP}}\big|\right]\\
&=\mathbb{E}\Bigg[\Bigg|\mathbb{E}\Bigg[\sum^K_{a=1}\frac{\left(\epol_t(a\mid X_t, \mathcal{S}_{t-1})\right)^2\big(Y_t(a) - \hat{f}_{t-1}(a, X_t)\big)^2}{\bpol(a\mid X_t)} + \left(\sum^K_{a=1}\epol_t(a\mid X_t, \mathcal{S}_{t-1})\hat{f}_{t-1}(a, X_t) - R(\epol_t)\right)^2+ \\
&\ \ \ \ \ \ \ \ \ 2\left(\sum^K_{a=1}\epol_t(a\mid X_t, \mathcal{S}_{t-1})f^*(a, X_t) - \sum^K_{a=1}\epol_t(a\mid X_t, \mathcal{S}_{t-1})\hat{f}_{t-1}(a, X_t)\right)\left(\sum^K_{a=1}\epol_t(a\mid X_t, \mathcal{S}_{t-1})\hat{f}_{t-1}(a, X_t) - R(\epol_t)\right) - \\
&\ \ \ \ \ \ \ \ \ \sum^{K}_{a=1}\frac{\big(\epol(a\mid X_t)\big)^2\big(Y_t(a)- f^*(a, X_t)\big)^2}{\bpol(a\mid X_t)} - \Big(\sum^{K}_{a=1}\epol(a\mid X_t)f^*(a, X_t) - R(\epol)\Big)^2\mid \mathcal{S}_{t-1}\Bigg]\Bigg|\Bigg]\\
&=\mathbb{E}\Bigg[\Bigg|\mathbb{E}\Bigg[\mathbb{E}\Bigg[\sum^K_{a=1}\frac{\left(\epol_t(a\mid X_t, \mathcal{S}_{t-1})\right)^2\big(Y_t(a) - \hat{f}_{t-1}(a, X_t)\big)^2}{\bpol(a\mid X_t)} + \left(\sum^K_{a=1}\epol_t(a\mid X_t, \mathcal{S}_{t-1})\hat{f}_{t-1}(a, X_t) - R(\epol_t)\right)^2+ \\
&\ \ \ \ \ \ \ \ \ 2\left(\sum^K_{a=1}\epol_t(a\mid X_t, \mathcal{S}_{t-1})f^*(a, X_t) - \sum^K_{a=1}\epol_t(a\mid X_t, \mathcal{S}_{t-1})\hat{f}_{t-1}(a, X_t)\right)\left(\sum^K_{a=1}\epol_t(a\mid X_t, \mathcal{S}_{t-1})\hat{f}_{t-1}(a, X_t) - R(\epol_t)\right) - \\
&\ \ \ \ \ \ \ \ \ \sum^{K}_{a=1}\frac{\big(\epol(a\mid X_t)\big)^2\big(Y_t(a)- f^*(a, X_t)\big)^2}{\bpol(a\mid X_t)} - \Big(\sum^{K}_{a=1}\epol(a\mid X_t)f^*(a, X_t) - R(\epol)\Big)^2\mid X_t, \mathcal{S}_{t-1}\Bigg] \mid \mathcal{S}_{t-1}\Bigg]\Bigg|\Bigg].
\end{align*}
Then, by using Jensen's inequality, 

\begin{align*}
&\mathbb{E}\left[\big|\mathbb{E}\big[Z^2_t\mid \mathcal{S}_{t-1}\big] - \sigma^2\big|\right]\\
&\leq \mathbb{E}\Bigg[\mathbb{E}\Bigg[\Bigg|\mathbb{E}\Bigg[\sum^K_{a=1}\frac{\left(\epol_t(a\mid X_t, \mathcal{S}_{t-1})\right)^2\big(Y_t(a) - \hat{f}_{t-1}(a, X_t)\big)^2}{\bpol(a\mid X_t)} + \left(\sum^K_{a=1}\epol_t(a\mid X_t, \mathcal{S}_{t-1})\hat{f}_{t-1}(a, X_t) - R(\epol_t)\right)^2+ \\
&\ \ \ \ \ \ \ \ \ 2\left(\sum^K_{a=1}\epol_t(a\mid X_t, \mathcal{S}_{t-1})f^*(a, X_t) - \sum^K_{a=1}\epol_t(a\mid X_t, \mathcal{S}_{t-1})\hat{f}_{t-1}(a, X_t)\right)\left(\sum^K_{a=1}\epol_t(a\mid X_t, \mathcal{S}_{t-1})\hat{f}_{t-1}(a, X_t) - R(\epol_t)\right) - \\
&\ \ \ \ \ \ \ \ \ \sum^{K}_{a=1}\frac{\big(\epol(a\mid X_t)\big)^2\big(Y_t(a)- f^*(a, X_t)\big)^2}{\bpol(a\mid X_t)} - \Big(\sum^{K}_{a=1}\epol(a\mid X_t)f^*(a, X_t) - R(\epol)\Big)^2\mid X_t, \mathcal{S}_{t-1}\Bigg] \Bigg| \mid \mathcal{S}_{t-1}\Bigg]\Bigg]\\
&= \mathbb{E}\Bigg[\Bigg|\mathbb{E}\Bigg[\sum^K_{a=1}\frac{\left(\epol_t(a\mid X_t, \mathcal{S}_{t-1})\right)^2\big(Y_t(a) - \hat{f}_{t-1}(a, X_t)\big)^2}{\bpol(a\mid X_t)} + \left(\sum^K_{a=1}\epol_t(a\mid X_t, \mathcal{S}_{t-1})\hat{f}_{t-1}(a, X_t) - R(\epol_t)\right)^2+ \\
&\ \ \ \ \ \ \ \ \ 2\left(\sum^K_{a=1}\epol_t(a\mid X_t, \mathcal{S}_{t-1})f^*(a, X_t) - \sum^K_{a=1}\epol_t(a\mid X_t, \mathcal{S}_{t-1})\hat{f}_{t-1}(a, X_t)\right)\left(\sum^K_{a=1}\epol_t(a\mid X_t, \mathcal{S}_{t-1})\hat{f}_{t-1}(a, X_t) - R(\epol_t)\right) - \\
&\ \ \ \ \ \ \ \ \ \sum^{K}_{a=1}\frac{\big(\epol(a\mid X_t)\big)^2\big(Y_t(a)- f^*(a, X_t)\big)^2}{\bpol(a\mid X_t)} - \Big(\sum^{K}_{a=1}\epol(a\mid X_t)f^*(a, X_t) - R(\epol)\Big)^2\mid X_t, \mathcal{S}_{t-1}\Bigg] \Bigg| \Bigg].
\end{align*}
Because $\hat f_{t-1}$ and $\epol_t$ are constructed from $\mathcal{S}_{t-1}$,

\begin{align*}
&\mathbb{E}\left[\big|\mathbb{E}\big[Z^2_t\mid \mathcal{S}_{t-1}\big] - \sigma^2\big|\right]\\
&\leq \mathbb{E}\Bigg[\Bigg|\sum^K_{a=1}\frac{\left(\epol_t(a\mid X_t, \mathcal{S}_{t-1})\right)^2\big(Y_t(a) - \hat{f}_{t-1}(a, X_t)\big)^2}{\bpol(a\mid X_t)} + \left(\sum^K_{a=1}\epol_t(a\mid X_t, \mathcal{S}_{t-1})\hat{f}_{t-1}(a, X_t) - R(\epol_t)\right)^2+ \\
&\ \ \ \ \ \ \ \ \ 2\left(\sum^K_{a=1}\epol_t(a\mid X_t, \mathcal{S}_{t-1})f^*(a, X_t) - \sum^K_{a=1}\epol_t(a\mid X_t, \mathcal{S}_{t-1})\hat{f}_{t-1}(a, X_t)\right)\left(\sum^K_{a=1}\epol_t(a\mid X_t, \mathcal{S}_{t-1})\hat{f}_{t-1}(a, X_t) - R(\epol_t)\right) - \\
&\ \ \ \ \ \ \ \ \ \sum^{K}_{a=1}\frac{\big(\epol(a\mid X_t)\big)^2\big(Y_t(a)- f^*(a, X_t)\big)^2}{\bpol(a\mid X_t)} - \Big(\sum^{K}_{a=1}\epol(a\mid X_t)f^*(a, X_t) - R(\epol)\Big)^2\mid X_t, \hat{f}_{t-1}, \epol_t\Bigg] \Bigg| \Bigg].
\end{align*}
From the results of Step~1, there exist $\tilde{C}_4$ and $\tilde{C}_5$ such that
\begin{align*}
&\mathbb{E}\left[\big|\mathbb{E}\big[Z^2_t\mid \mathcal{S}_{t-1}\big] - \sigma^2\big|\right]\\
&\leq \mathbb{E}\Bigg[\Bigg|\mathbb{E}\Bigg[\sum^K_{a=1}\frac{\left(\epol_t(a\mid X_t, \mathcal{S}_{t-1})\right)^2\big(Y_t(a) - \hat{f}_{t-1}(a, X_t)\big)^2}{\bpol(a\mid X_t)} + \left(\sum^K_{a=1}\epol_t(a\mid X_t, \mathcal{S}_{t-1})\hat{f}_{t-1}(a, X_t) - R(\epol_t)\right)^2+ \\
&\ \ \ \ \ \ \ \ \ 2\left(\sum^K_{a=1}\epol_t(a\mid X_t, \mathcal{S}_{t-1})f^*(a, X_t) - \sum^K_{a=1}\epol_t(a\mid X_t, \mathcal{S}_{t-1})\hat{f}_{t-1}(a, X_t)\right)\left(\sum^K_{a=1}\epol_t(a\mid X_t, \mathcal{S}_{t-1})\hat{f}_{t-1}(a, X_t) - R(\epol_t)\right) - \\
&\ \ \ \ \ \ \ \ \ \sum^{K}_{a=1}\frac{\big(\epol(a\mid X_t)\big)^2\big(Y_t(a)- f^*(a, X_t)\big)^2}{\bpol(a\mid X_t)} - \Big(\sum^{K}_{a=1}\epol(a\mid X_t)f^*(a, X_t) - R(\epol)\Big)^2\mid X_t, \hat{f}_{t-1}, \epol_t\Bigg] \Bigg| \Bigg]\\
&\leq \tilde{C}_4 \sum^1_{k=0}\mathbb{E}\left[\Big| \epol_t(a\mid X_t, \mathcal{S}_{t-1}) - \epol(a\mid X_t) \Big|\right] + \tilde{C}_5 \sum^K_{a=1}\mathbb{E}\left[\Big| \hat{f}_{t-1}(a, X_t) - f^*(a, X_t) \Big|\right].
\end{align*}
Then, from $L^r$ convergence theorem, by using pointwise convergence of $\epol_t$ and $\hat{f}_{t-1}$ and the boundedness of $z_t$, we have $\mathbb{E}\left[\big|\mathbb{E}\big[Z^2_t\mid \mathcal{S}_{t-1}\big] - \sigma^2\big|\right]\to 0$. Therefore,

\begin{align*}
&\mathbb{P}\left(\left|\frac{1}{T}\sum^T_{t=1}\mathbb{E}\big[Z^2_t\mid \mathcal{S}_{t-1}\big] - \sigma^2\right| \geq \varepsilon\right) \leq \frac{\frac{1}{T}\sum^T_{t=1}\mathbb{E}\left[\big|\mathbb{E}\big[Z^2_t\mid \mathcal{S}_{t-1}\big] - \sigma^2\big|\right]}{\varepsilon} \to 0.
\end{align*}
As a conclusion, 

\begin{align*}
&\frac{1}{T}\sum^T_{t=1}Z^2_t - \sigma^2_{\mathrm{AP}} = \frac{1}{T}\sum^T_{t=1}\big(Z^2_t - \mathbb{E}\left[Z^2_t\mid \mathcal{S}_{t-1}\big] + \mathbb{E}\big[Z^2_t\mid \mathcal{S}_{t-1}\big] - \sigma^2_{\mathrm{AP}}\right)\xrightarrow{\mathrm{p}} 0.
\end{align*}

\subsection*{Conclusion}
We can use CLT for an MDS. Hence, we have

\begin{align*} 
\sqrt{T}\left(\widehat{R}^{\mathrm{AP}}\left(\{\epol_t\}^T_{t=1}\right) - \frac{1}{T}\sum^T_{t=1}R(\epol_t)\right) \to \mathcal{N}\left(0, \sigma^2_{\mathrm{AP}}\right),
\end{align*}
where $\sigma^2_{\mathrm{AP}} = \mathbb{E}\left[\sum^{K}_{a=1}\frac{\big(\epol(a\mid X)\big)^2\nu^*\big(a, X\big)}{\bpol(a\mid X)} + \Big(\sum^{K}_{a=1}\epol(a\mid X)f^*(a, X) - R(\epol)\Big)^2\right]$.
\end{proof}

\section{OPE with Hypothesis Testing}
\label{appdx:hyp_test}
Here, we consider constructing evaluation probabilities by using $\mathcal{S}^{(1)}_{T^{(1)}}$ and evaluate them by using $\mathcal{S}^{(2)}_{T^{(2)}}$. Let us assume the number of evaluation probabilities is $M=2$, and the $L$ OPE estimators jointly follows the asymptotic distribution with the covariance matrix $\Sigma^{L}$, i.e., for a $L$ dimensional vector $I_L=(1\ 1\ \cdots\ 1)$ $\sqrt{T}\left(\widehat{\bm{R}}(\epol) - I_LR(\epol)\right)\xrightarrow{\mathrm{d}}\mathcal{N}(0, \Sigma)$, where $\widehat{\bm{R}}(\epol) = \left(\widehat{R}^1(\epol)\ \cdots\ \widehat{R}^L(\epol)\right)^\top$. Then, if $\Sigma$ is known, we can obtain an efficient estimator as $\widehat{R}^{\mathrm{efficient}}(\epol) = \left(I^\top \Sigma^{-1} I \right)^{-1}I^\top \Sigma^{-1} \widehat{\bm{R}}(\epol)$, which has the variance $\sigma^2_{\mathrm{efficient}} = \left(I^\top \Sigma^{-1} I \right)^{-1}$ \citep{GVK126800421,greene2003econometric}. Under these settings, we consider test the null hypothesis $R(\pi^{\mathrm{e}, (1)}) = R(\pi^{\mathrm{e}, (2)})$. By using an efficient OPE estimator $\widehat{R}^{\mathrm{efficient}}(\pi^{\mathrm{\es, (1)}}) - \widehat{R}^{\mathrm{efficient}}(\pi^{\mathrm{\es, (2)}})$, which has the asymptotic normality, we can conduct hypothesis testing. In general, we can estimate $R^{\mathrm{DVE}}(\pi^{\mathrm{\es, (1)}}, \pi^{\mathrm{\es, (2)}})$ by $\widehat{R}^{\mathrm{efficient}}(\pi^{\mathrm{\es, (1)}} - \pi^{\mathrm{\es, (2)}})$.

\begin{remark}[Cross hypothesis testing]
As the cross-validation, after the above process, we can construct evaluation probabilities by using $\mathcal{S}^{(1)}_{T^{(1)}}$ and test them by using $\mathcal{S}^{(2)}_{T^{(2)}}$. However, this additional hypothesis testing cases multiple hypothesis testing and should be avoided if possible. When conducting such a process, we can correct the confidence interval by multiple testing methods such as Bonferroni correction.
\end{remark}

\section{Other Discussions}
\label{appdx:det_discuss}

\subsection{Variance-regret trade-off}
There is a trade-off between regret and adaptability of OPE.  Let us consider the setting where there is no covariate, i.e., we can only observe $(A_t, Y_t)$. When it is possible to perform OPE for general evaluation policies, the behavior policy cannot achieve the best order of the expected regret. Let $\Theta$ be a set of all such problems satisfying Assumption~\ref{asm:overlap_reward}. For each $\theta \in \Theta$, let us define the expected regret as follows:

\begin{align*}
\mathrm{regret}_\theta \coloneqq \sum_{a \neq a^*}\Delta_{a}\mathbb{E}\left[\sum^T_{t=1}\mathbbm{1}[A_t=a]\right],
\end{align*}
where $\Delta_{a} 
\coloneqq \mathbb{E}\left[Y_t(a^*) - Y_t(a)\right]$ and $ a^* \in  \argmax_{a \in [K]} \mathbb{E}\left[Y_t(a)\right]$. 
The goal of the MAB problem is to minimize the expected regret.  We call an algorithm is {\it consistent} if for all $\theta \in \Theta$, $ \mathrm{regret}_\theta = o(T^\alpha)$ for all $\alpha \in (0,1)$.  Applying the OPE algorithm requires the additional assumption that, for all $t$ and $a\in[K]$, $0\leq \frac{\epol(a)}{\bpol_t(a)} \leq C_1$. (Assumption~\ref{asm:overlap_pol}). Therefore, if we want to conduct OPE using the samples generated from MAB algorithms, the MAB algorithms need to perform uniform exploration with some small constant $\gamma \in (0, 1)$. This will cause additional $\gamma T$ regret. 
The theoretical regret lower bound for any consistent algorithm for most of the Stochastic MAB problem is known to be $\Omega(\log T)$ and there exists an algorithm to achieve $\mathcal{O}(\log T)$ regret, asymptotically \cite{lai1985asymptotically}. It implies when it is possible to perform OPE for general evaluation policies, the behavior policy cannot achieve the best order of the expected regret. Thus, there is a trade-off between the regret minimization of behavior MAB policies and OPE evaluation. 

\paragraph{Importance of Asymptotic Normality:} The asymptotic normality is an important criterion for obtaining a confidence interval and convergence rate. For obtaining the asymptotic normality, we need several conditions for OPE estimators. For example, \citet{ChernozhukovVictor2018Dmlf} devised DDM, which guarantees the asymptotic normality when using nonparametric models for estimating $f^*$ under a mild condition. Although such techniques are incorporated in various OPE methods, it is also reported that such methods worsen the estimator in the sense of MSE between the true value and OPE estimators. Therefore, by giving up the asymptotic normality, we may increase the performance empirically.

\begin{table}[t]
\caption{Specification of datasets}
\label{Dataset}
\begin{center}
\scalebox{0.9}[0.9]{\begin{tabular}{cccc}
\hline
Dataset&the number of samples &Dimension &the number of classes\\
\hline
mnist & 60,000 &  780 & 10 \\
pendigits & 7,496 &  16 & 10\\
sensorless& 58,509 &  48 & 11 \\
connect-4& 67,557 &  126 & 3\\
\end{tabular}}
\end{center}
\end{table}

\section{Details of Experiments}
\label{appdx:exp}
In this section, we describe the details of experiments. The dataset description is shown in Table~\ref{Dataset}. We show the results using sensorless and connect-4 datasets in Tables~\ref{tbl:exp_result4}--\ref{tbl:exp_result6}. The other information is described as follows.

\paragraph{Evaluation policy:} All algorithms are implemented by scikit-learn, which is one of the most famous machine learning libraries in Python\footnote{\url{https://scikit-learn.org/stable/}}. We solve classification problems and regard the output with softmax function as the evaluation probability. For all methods, we use cross-validation for the hyper-parameter tuning. For the SVM with RBF kernel, we select the Kernel coefficient from a set $\{0.01, 0.1, 1\}$. For the random forest, we select the max depth from a set $\{5, 10, 15, 20\}$ and the number of estimators from a set $\{10, 50, 100\}$. For all algorithms, we applied L2 regularization with a parameter chosen from $\{0.01, 0.1, 1\}$. The cross-validation is $2$-fold.

\paragraph{Estimator of $f^*$:} In DM LR, we use a naive linear regression. For the other methods, we estimate $f^*$ by using the kernel ridge regression with the Kernel coefficient from a set $\{0.01, 0.1, 1\}$ and regularization parameter chosen from a set $\{0.01, 0.1, 1\}$.  

\paragraph{OPE estimators:} For AIPW, we apply $2$-fold DDM for guaranteeing the asymptotic normality.

\begin{table*}[t]
\caption{Experimental results of OPE2D with various OPE estimators using mnist and pendigits datasets. The best and worst methods are highlighted in red and blue, respectively.}
\label{tbl:exp_result4}
\scalebox{0.65}[0.65]{
\begin{tabular}{l|rr|rr|rr|rr|rr|rr|rr|rr}
\toprule
{sensorless} &  \multicolumn{2}{c|}{IPW} &        \multicolumn{2}{c|}{DM LR} &        \multicolumn{2}{c|}{DM KR} &        \multicolumn{2}{c|}{AIPW} &        \multicolumn{2}{c|}{MEAN} &       \multicolumn{2}{c|}{Minimax} &       \multicolumn{2}{c|}{Mix} &       \multicolumn{2}{c}{Maxmax} \\
\hline
$\alpha$ &        Mean &        SD &        Mean &        SD &        Mean &        SD &        Mean &        SD &        Mean &        SD &       Mean &        SD &       Mean &        SD &       Mean &        SD \\
\hline
0.7 &  0.00221 &  0.00652 &  \textcolor{blue}{0.02250} &  0.04192 &  0.00924 &  0.01374 &  \textcolor{red}{0.00209} &  0.00640 &  0.00620 &  0.01947 &  0.00862 &  0.02834 &  0.00871 &  0.02810 &  0.01831 &  0.03417 \\
0.4 &  \textcolor{red}{0.00108} &  0.00325 &  \textcolor{blue}{0.01807} &  0.03241 &  0.00362 &  0.00828 &  0.00182 &  0.00766 &  0.00655 &  0.02217 &  0.00810 &  0.01867 &  0.00801 &  0.01745 &  0.01084 &  0.02943 \\
0.0 &  \textcolor{red}{0.00000} &  0.00000 &  \textcolor{blue}{0.02865} &  0.05426 &  0.00128 &  0.00902 &  \textcolor{red}{0.00000} &  0.00000 &  0.00302 &  0.02279 &  0.01479 &  0.04042 &  0.01446 &  0.03705 &  0.01194 &  0.04022 \\
\bottomrule
\end{tabular}}

\scalebox{0.65}[0.65]{
\begin{tabular}{l|rr|rr|rr|rr|rr|rr|rr|rr}
\toprule
{connect-4}$\ $ &  \multicolumn{2}{c|}{IPW} &        \multicolumn{2}{c|}{DM LR} &        \multicolumn{2}{c|}{DM KR} &        \multicolumn{2}{c|}{AIPW} &        \multicolumn{2}{c|}{MEAN} &       \multicolumn{2}{c|}{Minimax} &       \multicolumn{2}{c|}{Mix} &       \multicolumn{2}{c}{Maxmax} \\
\hline
$\alpha$ &        Mean &        SD &        Mean &        SD &        Mean &        SD &        Mean &        SD &        Mean &        SD &       Mean &        SD &       Mean &        SD &       Mean &        SD \\
\hline
0.7 &  0.00111 &  0.00354 &  0.02257 &  0.02314 &  \textcolor{red}{0.00069} &  0.00193 &  0.00111 &  0.00323 &  0.02257 &  0.02314 &  0.01160 &  0.02007 &  \textcolor{blue}{0.27933} &  0.31967 &  0.01213 &  0.01965 \\
0.4 &  0.00169 &  0.00417 &  0.02054 &  0.02070 &  \textcolor{red}{0.00072} &  0.00202 &  0.00113 &  0.00383 &  0.02054 &  0.02070 &  0.01095 &  0.01711 &  \textcolor{blue}{0.22867} &  0.29530 &  0.01162 &  0.01835 \\
0.0 &  0.00174 &  0.00677 &  0.02161 &  0.02339 &  0.00087 &  0.00309 &  \textcolor{red}{0.00040} &  0.00178 &  0.02161 &  0.02339 &  0.00973 &  0.01743 &  \textcolor{blue}{0.18617} &  0.26070 &  0.01269 &  0.02163 \\
\bottomrule
\end{tabular}}

\caption{Experimental results of ISOPE with various OPE estimators using mnist and pendigits datasets. The best and worst methods are highlighted in red and blue, respectively.}
\label{tbl:exp_result5}
\scalebox{0.65}[0.65]{
\begin{tabular}{l|rr|rr|rr|rr|rr|rr|rr|rr}
\toprule
{sensorless} &  \multicolumn{2}{c|}{IPW} &        \multicolumn{2}{c|}{DM LR} &        \multicolumn{2}{c|}{DM KR} &        \multicolumn{2}{c|}{AIPW} &        \multicolumn{2}{c|}{MEAN} &       \multicolumn{2}{c|}{Minimax} &       \multicolumn{2}{c|}{Mix} &       \multicolumn{2}{c}{Maxmax} \\
\hline
$\alpha$ &        Mean &        SD &        Mean &        SD &        Mean &        SD &        Mean &        SD &        Mean &        SD &       Mean &        SD &       Mean &        SD &       Mean &        SD \\
\hline
0.7 &  0.00399 &  0.01230 &  \textcolor{blue}{0.01564} &  0.03957 &  \textcolor{red}{0.00036} &  0.00134 &  0.00379 &  0.01258 &  0.00550 &  0.02082 &  0.00542 &  0.02509 &  0.00644 &  0.02489 &  0.00957 &  0.02786 \\
0.4 &  0.00698 &  0.01331 &  \textcolor{blue}{0.01688} &  0.03630 &  \textcolor{red}{0.00116} &  0.00380 &  0.00432 &  0.00978 &  0.00527 &  0.01787 &  0.00279 &  0.00922 &  0.00402 &  0.01459 &  0.01592 &  0.03300 \\
0.0 &  \textcolor{red}{0.00000} &  0.00000 &  \textcolor{blue}{0.01921} &  0.05038 &  \textcolor{red}{0.00000} &  0.00000 &  \textcolor{red}{0.00000} &  0.00000 &  \textcolor{red}{0.00000} &  0.00000 &  0.01320 &  0.04256 &  0.01363 &  0.04230 &  0.00353 &  0.02480 \\
\bottomrule
\end{tabular}}

\scalebox{0.65}[0.65]{
\begin{tabular}{l|rr|rr|rr|rr|rr|rr|rr|rr}
\toprule
{connect-4}$\ $ &  \multicolumn{2}{c|}{IPW} &        \multicolumn{2}{c|}{DM LR} &        \multicolumn{2}{c|}{DM KR} &        \multicolumn{2}{c|}{AIPW} &        \multicolumn{2}{c|}{MEAN} &       \multicolumn{2}{c|}{Minimax} &       \multicolumn{2}{c|}{Mix} &       \multicolumn{2}{c}{Maxmax} \\
\hline
$\alpha$ &        Mean &        SD &        Mean &        SD &        Mean &        SD &        Mean &        SD &        Mean &        SD &       Mean &        SD &       Mean &        SD &       Mean &        SD \\
\hline
0.7 &  0.03619 &  0.02002 &  0.02581 &  0.02343 &  \textcolor{red}{0.00557} &  0.00884 &  0.02599 &  0.02291 &  0.02581 &  0.02343 &  0.01553 &  0.02071 &  \textcolor{blue}{0.24113} &  0.33316 &  0.03176 &  0.02267 \\
0.4 &  0.02822 &  0.02205 &  0.02189 &  0.02176 &  \textcolor{red}{0.00261} &  0.00655 &  0.02156 &  0.02231 &  0.02189 &  0.02176 &  0.01412 &  0.01900 &  \textcolor{blue}{0.25726} &  0.31583 &  0.02365 &  0.02304 \\
0.0 &  0.02613 &  0.02522 &  0.01922 &  0.02129 &  \textcolor{red}{0.00292} &  0.00619 &  0.01854 &  0.02291 &  0.01922 &  0.02129 &  0.00974 &  0.01611 &  \textcolor{blue}{0.19965} &  0.26632 &  0.02596 &  0.02362 \\
\bottomrule
\end{tabular}}

\caption{Experimental results of OPCV with various OPE estimators using mnist and pendigits datasets. The best and worst methods are highlighted in red and blue, respectively.}
\label{tbl:exp_result6}
\scalebox{0.65}[0.65]{
\begin{tabular}{l|rr|rr|rr|rr|rr|rr|rr|rr}
\toprule
{sensorless} &  \multicolumn{2}{c|}{IPW} &        \multicolumn{2}{c|}{DM LR} &        \multicolumn{2}{c|}{DM KR} &        \multicolumn{2}{c|}{AIPW} &        \multicolumn{2}{c|}{MEAN} &       \multicolumn{2}{c|}{Minimax} &       \multicolumn{2}{c|}{Mix} &       \multicolumn{2}{c}{Maxmax} \\
\hline
$\alpha$ &        Mean &        SD &        Mean &        SD &        Mean &        SD &        Mean &        SD &        Mean &        SD &       Mean &        SD &       Mean &        SD &       Mean &        SD \\
\hline
0.7 &  \textcolor{red}{0.01362} &  0.02798 &  0.01814 &  0.02833 &  0.01956 &  0.01624 &  0.01459 &  0.03088 &  0.01632 &  0.03128 &  0.01952 &  0.03225 &  \textcolor{blue}{0.02035} &  0.02914 &  0.01386 &  0.02542 \\
0.4 &  0.01913 &  0.02943 &  \textcolor{blue}{0.02450} &  0.03072 &  0.01810 &  0.02203 &  0.02249 &  0.03225 &  \textcolor{red}{0.01587} &  0.02487 &  0.01848 &  0.02390 &  0.02000 &  0.02299 &  0.02133 &  0.02910 \\
0.0 &  \textcolor{blue}{0.00290} &  0.02897 &  \textcolor{red}{0.00000} &  0.00000 &  \textcolor{red}{0.00000} &  0.00000 &  \textcolor{red}{0.00000} &  0.00000 &  \textcolor{red}{0.00000} &  0.00000 &  \textcolor{red}{0.00000} &  0.00000 & \textcolor{red}{0.00000} &  0.00000 &  \textcolor{blue}{0.00290} &  0.02897 \\
\bottomrule
\end{tabular}}

\scalebox{0.65}[0.65]{
\begin{tabular}{l|rr|rr|rr|rr|rr|rr|rr|rr}
\toprule
{connect-4}$\ $ &  \multicolumn{2}{c|}{IPW} &        \multicolumn{2}{c|}{DM LR} &        \multicolumn{2}{c|}{DM KR} &        \multicolumn{2}{c|}{AIPW} &        \multicolumn{2}{c|}{MEAN} &       \multicolumn{2}{c|}{Minimax} &       \multicolumn{2}{c|}{Mix} &       \multicolumn{2}{c}{Maxmax} \\
\hline
$\alpha$ &        Mean &        SD &        Mean &        SD &        Mean &        SD &        Mean &        SD &        Mean &        SD &       Mean &        SD &       Mean &        SD &       Mean &        SD \\
\hline
0.7 &  \textcolor{red}{0.00270} &  0.00450 &  0.01889 &  0.02271 &  0.00295 &  0.00459 &  0.00360 &  0.00660 &  0.01889 &  0.02271 &  0.00779 &  0.01488 &     \textcolor{blue}{0.23013} &      0.31443 &  0.01568 &  0.02167 \\
0.4 &  0.00837 &  0.00992 &  0.02394 &  0.02142 &  0.00836 &  0.01010 &  \textcolor{red}{0.00768} &  0.01061 &  0.02394 &  0.02142 &  0.01818 &  0.02002 &  \textcolor{blue}{0.27391} &  0.30736 &  0.01426 &  0.01659 \\
0.0 &  0.01630 &  0.02137 &  0.02636 &  0.02269 &  0.01704 &  0.02156 &  \textcolor{red}{0.01567} &  0.02045 &  0.02636 &  0.02269 &  0.02136 &  0.02231 &     \textcolor{blue}{0.25140} &      0.27160 &  0.01976 &  0.02244 \\
\bottomrule
\end{tabular}}
\end{table*}

\section{Proof of Theorem \ref{thm:game}}
\begin{proof}
First, $p^{\ast}$ is given as the follows:

\begin{align*}
  p^{\ast} = \argmax_{p \in \Delta^{M-1}} \min_{w \in \Delta^{L-1}} \sum_{j=1}^L w_j \hat{R}^j(\pi^{\mathrm{e}}_p),
\end{align*}
where $\pi^{\mathrm{e}}_p = \sum_{i=1}^M p_i \pi^{\mathrm{e}, (i)}$.
Since $\hat{R}^j(\pi) = \frac{1}{T} \sum_{t=1}^T\langle \pi(X_t), \Gamma_{j,t}\rangle$, we have:

\begin{align*}
  \sum_{j=1}^L w_j \hat{R}^j(\pi^{\mathrm{e}}_p) &= \sum_{j=1}^L w_j \left( \frac{1}{T}\sum_{t=1}^T \left\langle \sum_{i=1}^M p_i \pi^{\mathrm{e}, (i)}(X_t), \Gamma_{j,t}\right\rangle \right) \\
  &= \sum_{j=1}^L w_j \left( \frac{1}{T}\sum_{t=1}^T \sum_{i=1}^M p_i \left\langle \pi^{\mathrm{e}, (i)}(X_t), \Gamma_{j,t}\right\rangle \right) \\
  &= \sum_{j=1}^L w_j \sum_{i=1}^M p_i \left( \frac{1}{T}\sum_{t=1}^T \left\langle \pi^{\mathrm{e}, (i)}(X_t), \Gamma_{j,t}\right\rangle \right) \\
  &= \sum_{j=1}^L w_j \sum_{i=1}^M p_i \hat{R}^j(\pi^{\mathrm{e}, (i)}) \\
  &= \sum_{i=1}^M \sum_{j=1}^L p_i w_j \hat{R}^j(\pi^{\mathrm{e}, (i)}).
\end{align*}
  Therefore, we can rewrite $p^{\ast}$ as follows:
  
\begin{align*}
  p^{\ast} = \argmax_{p \in \Delta^{M-1}} \min_{w \in \Delta^{L-1}} \sum_{i=1}^M \sum_{j=1}^L p_i w_j \hat{R}^j(\pi^{\mathrm{e}, (i)}).
\end{align*}
This equation implies that $p^{\ast}$ is a Nash equilibrium strategy in the two-player zero-sum normal-form game with payoff matrix $(C_{ij})$ where $C_{ij} = \hat{R}^j(\pi^{\mathrm{e}, (i)})$.
We can rewrite $p^{\ast}$ as the solution of the following linear problem:

\begin{align*}
  &\max_{z, p} ~z \\
  \mathrm{s.t.} ~ & p^T C \geq z \mathbf{1}^T \\
  & p^T \mathbf{1} = 1 \\
  & \forall i, p_i \geq 0.
\end{align*}

\end{proof}

\end{document}